\documentclass[11pt]{article}
\usepackage{misc/acl}

\usepackage{times}
\usepackage{latexsym}
\usepackage[T1]{fontenc}
\usepackage[utf8]{inputenc}
\usepackage{microtype}
\usepackage{inconsolata}
\usepackage{graphicx}
\usepackage{tikz}
\usepackage[table,x11names]{xcolor}
\usepackage{amsmath,amssymb}
\usepackage{cleveref}
\usepackage{algorithm}
\usepackage{algorithmicx}
\usepackage[noend]{algpseudocode}
\usepackage{caption}
\usepackage{enumitem}
\setlist{noitemsep,left=0mm,topsep=1mm}
\usepackage{booktabs}
\usepackage{multirow}
\usepackage{array}
\usepackage{makecell}
\usepackage{setspace} %

\let\svthefootnote\thefootnote
\newcommand\blankfootnote[1]{%
  \let\thefootnote\relax\footnotetext{#1}%
  \let\thefootnote\svthefootnote%
}

\title{Generating Difficult-to-Translate Texts}

\author{
Vilém Zouhar$^{\bigstar}$
\quad Wenda Xu$^{\blacksquare}$
\quad Parker Riley$^{\blacksquare}$
\quad Juraj Juraska$^{\blacksquare}$ \\
\bf Mara Finkelstein$^{\blacksquare}$
\quad Markus Freitag$^{\blacksquare}$
\quad Daniel Deutsch$^{\blacksquare}$\\[0.2em]
$^{\blacksquare}$Google \quad $^{\bigstar}$ETH Zurich}

\begin{document}
\maketitle

\blankfootnote{$^{\bigstar}$Work done during Google Internship.\\\null\hspace{7mm} Correspondence: \href{mailto:vzouhar@ethz.ch}{vzouhar@ethz.ch}}

\begin{abstract}
Machine translation benchmarks sourced from the real world are quickly obsoleted, due to most examples being easy for state-of-the-art translation models.
This limits the benchmark's ability to distinguish which model is better or to reveal models' weaknesses.
Current methods for creating difficult test cases, such as subsampling or from-scratch synthesis, either fall short of identifying difficult examples or suffer from a lack of diversity and naturalness.
Inspired by the iterative process of human experts probing for model failures, we propose MT-breaker, a method where a large language model iteratively refines a source text to increase its translation difficulty.
The LLM iteratively queries a target machine translation model to guide its generation of difficult examples.
Our approach generates examples that are more challenging for the target MT model while preserving the diversity of natural texts.
While the examples are tailored to a particular machine translation model during the generation, the difficulty also transfers to other models and languages.
\end{abstract}

\section{Introduction}

Test sets need to be difficult in order to be informative.
If an NLP model performs flawlessly on a test set, then we learn very little about the model's shortcomings.
Or, if we compare two NLP models that perform near-perfectly on a test set, we do not know which one is better.
For machine translation, current models are able to translate even recent benchmarks with only a few mistakes across several language directions \citep{kocmi-etal-2024-findings}.
Attempts have been made to automatically select a subset of existing datasets that still pose challenges to the models \citep{proietti2025estimatingmachinetranslationdifficulty}.
However, these methods fall short due to the absence of a sufficient number of naturally occurring challenging examples.
Another line of work makes use of LLMs to create difficult examples from scratch in a zero-shot manner \citep{pombal2025zero}, which we show is inadequate due to its lack of difficulty and diversity.

\begin{figure}[t]
    \centering
    \includegraphics[width=1\linewidth]{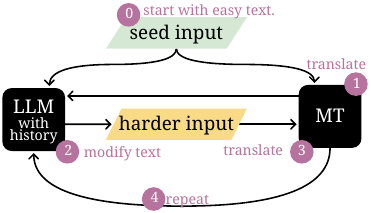}
    \caption{High-level overview of our interactive difficulty-to-translate text generation pipeline. The process is repeated for a fixed number of iterations.}
    \label{fig:highlevel}
\end{figure}

A human practitioner who is trying to assess the quality of a model approaches this task differently: not by selecting difficult examples or synthesizing them from scratch (finding based on interviews in \Cref{sec:human_qualitative}).
Instead, the practitioner iteratively tries to find weaknesses in the model.
They might start with some random inputs and then, based on the model's response, modify these inputs to cause the model to provide a faulty output.
This way, the practitioner arrives at difficult examples tailored to a particular model that reveal its shortcomings.
These insights can be then used to diagnose and improve the model.

\definecolor{mypurple}{HTML}{B5739D}
\newcommand{\purplecircle}[1]{%
  \tikz[baseline=(char.base)]{
    \node[
      shape=circle,          
      fill=mypurple,         
      text=white,            
      inner sep=1pt,         
    ] (char) {#1};
  }%
}

\begin{table*}[t]
\renewcommand{\arraystretch}{1.5}
\fontsize{8}{8}\selectfont
\begin{tabular}{ p{4.5cm}p{4.8cm}p{5.1cm} } \toprule
\bf\small Zeroshot (min) & \bf\small Seeds & \bf\small MTbreaker (seeded)\\
It is what it is. & Another one has been found! & Another rabbit hole of lies was found!\\
The quick brown fox jumps over the lazy dog. & I quit FB, removed all Meta and Pinterest tracker from my website and killed my private WhatsApp account. & I've quit FB and killed the private WhatsApp account to get rid of Meta's and Pinterest's gross ad-tracking tools that I hated with a fiery passion.\\
He bade them farewell and then he bade them all to hell. & Going back up tomorrow and we're doing stalls and slow flight. & Going back up tomorrow; we're doing stalls and spins, and then some unusual attitudes.\\
It is what it is. & Heheh not one but three! & Heheh not one, two, but three!!!11!oneeleven\\
\bottomrule\end{tabular}

\vspace{-3mm}
\caption{Example of generated sources in English by Zeroshot, Seeds, and MT-breaker (seeded). Examples were selected to be short. See Appendix \Cref{tab:10-examples_all} for more examples.}
\label{tab:10-examples_small}

\vspace{-1mm}
\end{table*}

In this work we propose MT-breaker, which uses an LLM to generate difficult-to-translate examples that mimic the process of the human practitioner (see \Cref{fig:highlevel}).
We start with a seed text \purplecircle{0} that the machine translation model translates \purplecircle{1}.
The text and its translation is then passed to an LLM which modifies it \purplecircle{2} and passes it to the MT model again \purplecircle{3}.
The translation is again passed back to the LLM model with chat-like history and the whole process repeats \purplecircle{4}.

We show that this approach maintains the diversity and naturalness of the source examples, while vastly increasing the difficulty of the generated examples.
Despite the approach relying on a quality estimation metric, the same findings are confirmed by an expert MQM human-evaluation study.
The examples are tailored to be difficult for a particular machine translation model, though we also show that the difficulty also transfers to other models and languages.
The generated dataset can be used to find weaknesses of a particular model, which can, in turn, be used for hillclimbing during model development.

\begin{figure}[t]
\hrule\vspace{1mm}
\small
\begin{algorithmic}[1]
\Statex \hspace{-5.5mm} \textbf{MT-breaker}(seed text $s_0$, steps $N$):
\For{$i = 1 \ldots N$}
\State $t_{i-1} \gets \mathrm{MT}(s_{i-1})$
\Comment{Translate}
\State $s_{i} \gets \mathrm{LLM}_\mathrm{step}(\{s_k\}_{k=0}^{i-1}, \{t\}_{k=0}^{i-1})$
\Comment{Generate next}
\EndFor
\State \textbf{return} $\arg\min \mathrm{QE}(\{s_k\}_{k=0}^{N}, \{t\}_{k=0}^{N})$
\Comment{Pick difficult}
\end{algorithmic}
\vspace{1mm}
\hrule

\medskip
\captionof{algorithm}{Given a source text, the LLM receives the MT model's translation which informs how to next change the text. The $\mathrm{LLM}_\mathrm{step}$ can be instructed to not stray too far from the original meaning or to preserve naturalness. At the end, we pick the source text from $s$ that led to the worst translation.}
\label{alg:mt_breaker}

\vspace{-5mm}
\end{figure}

\section{Methods}

We now describe the various approaches of generating difficult-to-translate texts.
The baseline, which we refer to as \textbf{seeds}, consists of simply taking existing texts from a dataset; i.e. not creating new ones.

\paragraph{Zero-shot benchmarking.}
This method simply prompts a large language model to generate an example to be part of a benchmark.
This generation can be steered towards particular topics, domains, and also difficulty.
This method is a simplified version of \citet{pombal2025zero}.
We introduce a variant ``(history)'', where we include the history of previously generated texts in the prompt such that they are not repeated.

\paragraph{MT-breaker.}
Our method is shown in \Cref{fig:highlevel} and \Cref{alg:mt_breaker}.
It is based on mimicking a human expert that tries to find an input that triggers an error in the machine translation model, which is a strategy motivated by qualitative interviews with human experts (\Cref{sec:human_qualitative}).
We start with a piece of text, called a seed, which is translated by a machine translation model.
This, together with the seed is the input to $\mathrm{LLM}_\mathrm{step}$, which modifies the seed.
This is repeated for a certain number of steps, e.g. 10.
At the end, we compute the quality estimation scores for all steps and select the most difficult source.
A quality estimation metric is a model that takes in the source and its translation and outputs an scalar assessment of the translation's quality \citep{freitag-etal-2024-llms}.
We also consider MT-breaker without the seed text ``(seedless)'', and a version which has the $\mathrm{LLM}_\mathrm{step}$ seeing the intermediate quality estimation scores ``(seeded+qe)''.
The prompts for $\mathrm{LLM}_\mathrm{step}$  in all variants are shown in \Cref{sec:prompts}.

\begin{figure*}[t]
    \centering
    \includegraphics[width=1\linewidth]{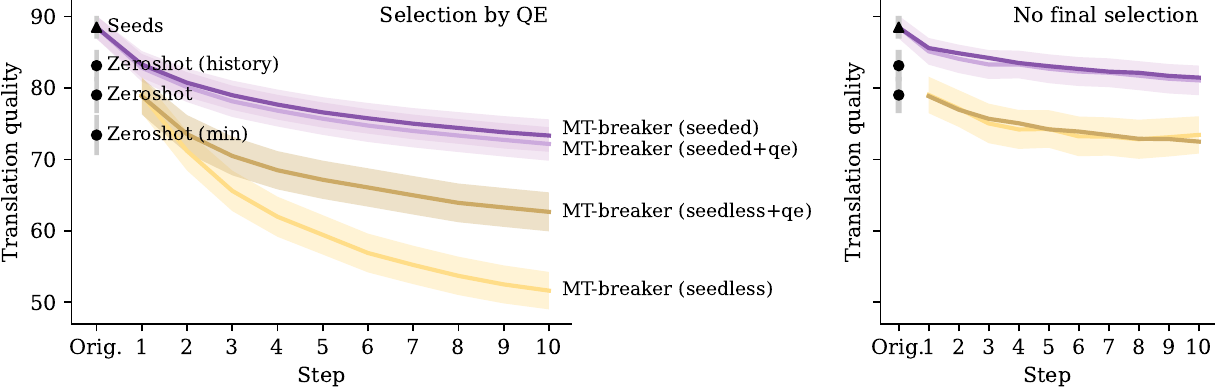}
    \caption{
    Development of difficulty with each iteration.
    Seeds require no iterations, Zeroshot and Zeroshot (history) require one LLM step, and the rest requires 10 steps.
    Right plot shows running hypotheses at a given step without final selection by quality estimation.
    This does not include Zeroshot (min) which is simply Zeroshot with selection.
    Values averaged across all models and languages.
    Shaded areas show 90\% t-test confidence intervals averaged across all models and languages.
    }
    \label{fig:01-difficulty_history}
\end{figure*}

\medskip
\section{Experiments and Results}
\label{sec:experiments}

We now describe experiments in which we compare and analyze the above listed methods for generating difficult-to-translate texts.

\paragraph{Setup.}
For brevity and computational feasibility, we consider English$\rightarrow$\{Czech, German, Chinese, Vietnamese, Polish\} language directions.
For seeds, we use 100 randomly sampled English sources from WMT 2024 \citep{kocmi-etal-2024-findings}.
We use the following publicly available machine translation models:
\begin{itemize}
\item Google Translate\footnote{
\href{https://translate.google.com}{translate.google.com}, accessed August 2025.
}
\item Gemini 2.5-flash \citep{gemini_team_2025_gemini2_5}
\item Gemini 2.0-flash \citep{gemini_team_2025_gemini2_5}
\item Gemma 3-27b-it \citep{gemma_team_2025_gemma3}
\end{itemize}

\noindent
For the LLM\textsubscript{step} we use Gemini 2.5-pro \citep{gemini_team_2025_gemini2_5} in the main paper, though explore others in \Cref{sec:experiments_transfer_llms}.
For measuring the quality of the translations, we use the average of MetricX-24-QE \citep{juraska-etal-2024-metricx} and prompted LLM-as-QE (see \Cref{sec:prompts} for details).
We also include a source-only version of MetricX, that only sees the source texts and no translation, which serves as a proxy for general difficulty (see \Cref{sec:srcqe_metricx}).
All three are scaled from 0 (poor translation, high difficulty) to 100 (perfect translation, low difficulty).
Same as \citet{proietti2025estimatingmachinetranslationdifficulty}, the quantity of ``difficulty'' has an opposite direction to quality estimation (100-QE).
While we use the term ``difficulty'', we include the original quality estimation scores in tables and figures for simplicity.

We compare the following methods for obtaining a difficult test set:
\begin{itemize}
\item ``Seed'' segments from WMT
\item ``Zeroshot'' benchmarking
\item ``Zeroshot (history)'' benchmarking with history
\item ``Zeroshot (min)'' benchmarking with selection\footnote{
For a fair comparison, we also include a version of Zeroshot that is run 10x (same number of steps as MT-breaker) out of which (per batch of 10) we select the most difficult sources with quality estimation in the same manner.
}
\item ``MT-breaker (seeded)'' with seeds
\item ``MT-breaker (seedless)'' without seeds
\item ``MT-breaker (seeded+qe)'' with seeds and with quality estimation
\item ``MT-breaker (seedless+qe)'' without seeds and with quality estimation
\end{itemize}
For a fair comparison, we instruct the generation of the Zeroshot benchmarking and MT-breaker without seeds methods to have approximately the same number of words as the seeds used by the other methods.

See generated examples of Zeroshot (min) and MT-breaker (seeded) in \Cref{tab:10-examples_small} and Appendix \Cref{tab:10-examples_all,tab:10-examples_history_encs,tab:10-examples_history_ende}.

\subsection{Increasing Difficulty}
\label{sec:experiments_difficulty}

As shown in \Cref{fig:01-difficulty_history}, the translation quality of texts generated by MT-breaker methods decreases with each step but begins to plateau around step 10.
Interestingly, having access to the quality estimation during the breaking process does not help the MT-breaker (seeded) and worsens MT-breaker (seedless).
We explain this through different variances.
The per-step averages (\Cref{fig:01-difficulty_history} right) are the same for the seedless version.
The reason for the lower minimum (shown cumulatively in the lines in \Cref{fig:01-difficulty_history} left) is that the version with quality estimation has lower variance ($\sigma^2$=297) than the version without it ($\sigma^2$=325).
This is likely caused by the model using most of the 10 steps to continue the breakage with the text with currently lowest quality estimation score, which can be a dead-end.
The minimum of a variable with higher variance is lower than the minimum of a variable with lower variance, even when the means are the same.

Methods that use seeds for the initial source (Seeds, MT-breaker seeded) lead to lower difficulty than those without such restrictions (Zeroshot, MT-breaker seedless).
In the next section, we show that this is part of a diversity-difficulty tradeoff.

\begin{table*}[t]
\centering
\small
\begin{tabular}{lcccccccc}
\toprule
& \multicolumn{1}{m{1.2cm}}{\centering\bfseries Seed} & \multicolumn{1}{m{1.2cm}}{\centering\bfseries Zeroshot} & \multicolumn{1}{m{1.2cm}}{\centering\bfseries Zeroshot} & \multicolumn{1}{m{1.2cm}}{\centering\bfseries Zeroshot} & \multicolumn{1}{m{1.2cm}}{\centering\bfseries MT-breaker} & \multicolumn{1}{m{1.2cm}}{\centering\bfseries MT-breaker} & \multicolumn{1}{m{1.2cm}}{\centering\bfseries MT-breaker} & \multicolumn{1}{m{1.2cm}}{\centering\bfseries MT-breaker} \\
& & & \multicolumn{1}{m{1.2cm}}{\centering\scriptsize(min)} & \multicolumn{1}{m{1.2cm}}{\centering\scriptsize(history)} & \multicolumn{1}{m{1.2cm}}{\centering\scriptsize(seedless)} & \multicolumn{1}{m{1.2cm}}{\centering\scriptsize(seeded)} & \multicolumn{1}{m{1.2cm}}{\centering\scriptsize(seedless+QE)} & \multicolumn{1}{m{1.2cm}}{\centering\scriptsize(seeded+QE)} \\
\midrule
\multicolumn{9}{l}{\it Diversity (higher is better)} \\
Diversity (embd) & \cellcolor{SeaGreen3!95}0.33 & \cellcolor{SeaGreen3!10}0.09 & \cellcolor{SeaGreen3!15}0.10 & \cellcolor{SeaGreen3!89}0.31 & \cellcolor{SeaGreen3!35}0.16 & \cellcolor{SeaGreen3!100}0.34 & \cellcolor{SeaGreen3!22}0.12 & \cellcolor{SeaGreen3!100}0.34 \\
Diversity (chrF) & \cellcolor{SeaGreen3!98}0.76 & \cellcolor{SeaGreen3!10}0.24 & \cellcolor{SeaGreen3!18}0.29 & \cellcolor{SeaGreen3!92}0.73 & \cellcolor{SeaGreen3!50}0.46 & \cellcolor{SeaGreen3!100}0.77 & \cellcolor{SeaGreen3!35}0.38 & \cellcolor{SeaGreen3!99}0.77 \\
Diversity (topics) & \cellcolor{SeaGreen3!89}247 & \cellcolor{SeaGreen3!10}126 & \cellcolor{SeaGreen3!14}131 & \cellcolor{SeaGreen3!73}225 & \cellcolor{SeaGreen3!25}148 & \cellcolor{SeaGreen3!99}262 & \cellcolor{SeaGreen3!24}147 & \cellcolor{SeaGreen3!100}262 \\
Diversity (errors) & \cellcolor{SeaGreen3!88}23.50 & \cellcolor{SeaGreen3!14}20.80 & \cellcolor{SeaGreen3!51}22.00 & \cellcolor{SeaGreen3!10}20.65 & \cellcolor{SeaGreen3!79}22.95 & \cellcolor{SeaGreen3!97}23.80 & \cellcolor{SeaGreen3!70}22.65 & \cellcolor{SeaGreen3!100}23.90 \\
Diversity (words) & \cellcolor{SeaGreen3!46}1566 & \cellcolor{SeaGreen3!10}1121 & \cellcolor{SeaGreen3!15}1177 & \cellcolor{SeaGreen3!39}1508 & \cellcolor{SeaGreen3!42}1532 & \cellcolor{SeaGreen3!98}2082 & \cellcolor{SeaGreen3!41}1527 & \cellcolor{SeaGreen3!100}2091 \\
\midrule
\multicolumn{9}{l}{\it Complexity \& Style} \\
Grammaticality & \cellcolor{Snow3!46}92.28 & \cellcolor{Snow3!76}96.16 & \cellcolor{Snow3!73}95.67 & \cellcolor{Snow3!90}97.91 & \cellcolor{Snow3!61}94.15 & \cellcolor{Snow3!10}87.65 & \cellcolor{Snow3!46}92.27 & \cellcolor{Snow3!12}87.93 \\
Naturalness & \cellcolor{Snow3!90}92.91 & \cellcolor{Snow3!42}57.49 & \cellcolor{Snow3!39}54.15 & \cellcolor{Snow3!65}73.89 & \cellcolor{Snow3!10}29.80 & \cellcolor{Snow3!79}82.71 & \cellcolor{Snow3!20}41.99 & \cellcolor{Snow3!80}83.31 \\
Word Rarity & \cellcolor{Snow3!10}17.67 & \cellcolor{Snow3!71}46.24 & \cellcolor{Snow3!71}46.14 & \cellcolor{Snow3!33}28.45 & \cellcolor{Snow3!90}54.80 & \cellcolor{Snow3!49}35.69 & \cellcolor{Snow3!86}53.23 & \cellcolor{Snow3!46}34.77 \\
Syntax Complexity & \cellcolor{Snow3!10}39.95 & \cellcolor{Snow3!51}55.44 & \cellcolor{Snow3!53}56.28 & \cellcolor{Snow3!56}57.12 & \cellcolor{Snow3!90}70.38 & \cellcolor{Snow3!39}51.21 & \cellcolor{Snow3!77}65.78 & \cellcolor{Snow3!37}50.23 \\
Avg. Word Count & \cellcolor{Snow3!33}33.25 & \cellcolor{Snow3!10}31.21 & \cellcolor{Snow3!10}31.25 & \cellcolor{Snow3!34}33.34 & \cellcolor{Snow3!11}31.32 & \cellcolor{Snow3!84}39.62 & \cellcolor{Snow3!69}37.67 & \cellcolor{Snow3!90}40.18 \\
Avg. Word Length & \cellcolor{Snow3!10}4.69 & \cellcolor{Snow3!61}5.44 & \cellcolor{Snow3!60}5.43 & \cellcolor{Snow3!26}4.93 & \cellcolor{Snow3!90}5.90 & \cellcolor{Snow3!35}5.07 & \cellcolor{Snow3!74}5.66 & \cellcolor{Snow3!31}5.01 \\
\midrule
\multicolumn{9}{l}{\it Quality Estimation (lower is better)} \\
QE (Gemini) & \cellcolor{Firebrick3!10}89.78 & \cellcolor{Firebrick3!23}81.94 & \cellcolor{Firebrick3!37}73.94 & \cellcolor{Firebrick3!21}83.53 & \cellcolor{Firebrick3!85}45.35 & \cellcolor{Firebrick3!48}67.52 & \cellcolor{Firebrick3!60}59.97 & \cellcolor{Firebrick3!45}69.23 \\
QE (MetricX) & \cellcolor{Firebrick3!10}87.25 & \cellcolor{Firebrick3!38}76.09 & \cellcolor{Firebrick3!44}72.87 & \cellcolor{Firebrick3!18}82.73 & \cellcolor{Firebrick3!85}57.88 & \cellcolor{Firebrick3!36}76.81 & \cellcolor{Firebrick3!63}65.33 & \cellcolor{Firebrick3!34}77.45 \\
SRCQE (MetricX) & \cellcolor{Firebrick3!10}94.47 & \cellcolor{Firebrick3!57}91.69 & \cellcolor{Firebrick3!60}91.55 & \cellcolor{Firebrick3!27}93.29 & \cellcolor{Firebrick3!82}90.54 & \cellcolor{Firebrick3!51}91.96 & \cellcolor{Firebrick3!85}90.12 & \cellcolor{Firebrick3!50}92.01 \\
\bottomrule
\end{tabular}

\caption{
Quantitative automatic evaluation of difficult data generation approaches.
}
\label{tab:01-data_quality}
\end{table*}

\begin{figure}[t]
    \centering
    \includegraphics[width=1\linewidth]{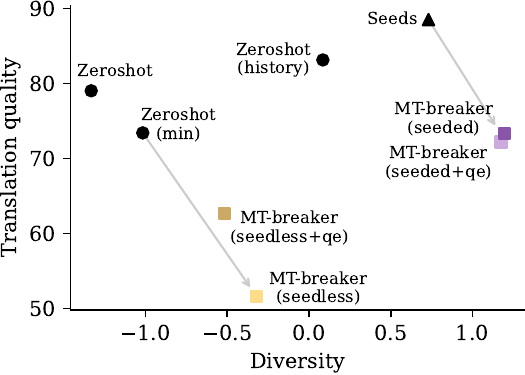}
    \caption{Comparison of difficulty and diversity of difficult-to-translate text generation methods. Diversity is z-normalized across each diversity measure and averaged.
    Gray arrows signify most fair comparison.
    }
    \label{fig:01-difficulty_pareto}
\end{figure}

\begin{table*}[t]
\small
\centering
\begin{tabular}{@{\hspace{1mm}}c@{\hspace{1mm}}l *{4}{>{\centering\arraybackslash}p{8.5mm}}}
    \toprule
    & & \multicolumn{4}{c}{Target model (translation quality)} \\
    &&
    \rotatebox[origin=c]{90}{Translate} &
    \rotatebox[origin=c]{90}{Gemma3} &
    \rotatebox[origin=c]{90}{Gemini2.0} &
    \rotatebox[origin=c]{90}{Gemini2.5}\\
    \cmidrule{3-6}
    \multirow{4}{*}{\rotatebox[origin=c]{90}{\parbox{1.5cm}{\centering\bf MT-break.\\(seedless)}}}
    & Gemma 3    & \cellcolor{Firebrick3!67}48.00 & \cellcolor{Firebrick3!61}57.31 & \cellcolor{Firebrick3!58}59.77 & \cellcolor{Firebrick3!56}58.99 \\
    & Gemini 2.0 & \cellcolor{Firebrick3!85}35.92 & \cellcolor{Firebrick3!85}42.36 & \cellcolor{Firebrick3!85}42.97 & \cellcolor{Firebrick3!85}39.79 \\
    & Gemini 2.5 & \cellcolor{Firebrick3!60}52.56 & \cellcolor{Firebrick3!56}60.51 & \cellcolor{Firebrick3!53}63.12 & \cellcolor{Firebrick3!51}62.18 \\
    & Gemini 2.5 pro  & \cellcolor{Firebrick3!73}44.13 & \cellcolor{Firebrick3!69}52.45 & \cellcolor{Firebrick3!64}55.91 & \cellcolor{Firebrick3!63}53.98 \\
    \midrule
    \multirow{4}{*}{\rotatebox[origin=c]{90}{\parbox{1.5cm}{\centering\bf Zeroshot\\(min)}}}
    & Gemma-3    & \cellcolor{Firebrick3!59}53.00 & \cellcolor{Firebrick3!44}68.28 & \cellcolor{Firebrick3!48}66.31 & \cellcolor{Firebrick3!44}66.75 \\
    & Gemini 2.0 & \cellcolor{Firebrick3!71}45.15 & \cellcolor{Firebrick3!66}54.34 & \cellcolor{Firebrick3!71}51.63 & \cellcolor{Firebrick3!61}55.67 \\
    & Gemini 2.5 & \cellcolor{Firebrick3!54}56.68 & \cellcolor{Firebrick3!51}63.62 & \cellcolor{Firebrick3!49}65.81 & \cellcolor{Firebrick3!48}64.31 \\
    & Gemini 2.5 pro  & \cellcolor{Firebrick3!31}71.78 & \cellcolor{Firebrick3!37}72.63 & \cellcolor{Firebrick3!35}74.59 & \cellcolor{Firebrick3!32}74.61 \\
    \midrule
    \multirow{4}{*}{\rotatebox[origin=c]{90}{\parbox{1.5cm}{\centering\bf MT-break.\\(seeded)}}}
    & Gemma-3    & \cellcolor{Firebrick3!58}53.69 & \cellcolor{Firebrick3!52}63.22 & \cellcolor{Firebrick3!50}65.05 & \cellcolor{Firebrick3!49}63.68 \\
    & Gemini 2.0 & \cellcolor{Firebrick3!76}41.67 & \cellcolor{Firebrick3!69}52.59 & \cellcolor{Firebrick3!68}53.56 & \cellcolor{Firebrick3!66}52.46 \\
    & Gemini 2.5 & \cellcolor{Firebrick3!49}59.86 & \cellcolor{Firebrick3!43}68.46 & \cellcolor{Firebrick3!37}73.11 & \cellcolor{Firebrick3!40}69.77 \\
    & Gemini 2.5 pro & \cellcolor{Firebrick3!41}65.01 & \cellcolor{Firebrick3!34}74.09 & \cellcolor{Firebrick3!35}74.47 & \cellcolor{Firebrick3!31}75.09 \\
    \midrule
    & \bf Seeds & \cellcolor{Firebrick3!10}85.63 & \cellcolor{Firebrick3!10}89.26 & \cellcolor{Firebrick3!10}89.94 & \cellcolor{Firebrick3!10}89.23 \\
    \bottomrule
\end{tabular}
\begin{tabular}{*{4}{>{\centering\arraybackslash}p{8.5mm}}}
    \toprule
    \multicolumn{4}{c}{Target model (diversity)} \\
    \rotatebox[origin=c]{90}{Translate} &
    \rotatebox[origin=c]{90}{Gemma3} &
    \rotatebox[origin=c]{90}{Gemini2.0} &
    \rotatebox[origin=c]{90}{Gemini2.5} \\
    \cmidrule{1-4}
    \cellcolor{SeaGreen3!55}\phantom{-}0.05 & \cellcolor{SeaGreen3!59}\phantom{-}0.19 & \cellcolor{SeaGreen3!64}\phantom{-}0.16 & \cellcolor{SeaGreen3!58}\phantom{-}0.06 \\
    \cellcolor{SeaGreen3!26}-1.03 & \cellcolor{SeaGreen3!34}-0.72 & \cellcolor{SeaGreen3!32}-0.95 & \cellcolor{SeaGreen3!24}-1.09 \\
    \cellcolor{SeaGreen3!66}\phantom{-}0.48 & \cellcolor{SeaGreen3!71}\phantom{-}0.64 & \cellcolor{SeaGreen3!75}\phantom{-}0.57 & \cellcolor{SeaGreen3!76}\phantom{-}0.65 \\
    \cellcolor{SeaGreen3!34}-0.72 & \cellcolor{SeaGreen3!45}-0.31 & \cellcolor{SeaGreen3!50}-0.30 & \cellcolor{SeaGreen3!44}-0.41 \\
    \midrule
    \cellcolor{SeaGreen3!27}-1.00 & \cellcolor{SeaGreen3!29}-0.89 & \cellcolor{SeaGreen3!27}-1.11 & \cellcolor{SeaGreen3!20}-1.23 \\
    \cellcolor{SeaGreen3!10}-1.62 & \cellcolor{SeaGreen3!10}-1.60 & \cellcolor{SeaGreen3!10}-1.72 & \cellcolor{SeaGreen3!10}-1.55 \\
    \cellcolor{SeaGreen3!54}\phantom{-}0.02 & \cellcolor{SeaGreen3!58}\phantom{-}0.18 & \cellcolor{SeaGreen3!55}-0.13 & \cellcolor{SeaGreen3!59}\phantom{-}0.08 \\
    \cellcolor{SeaGreen3!26}-1.03 & \cellcolor{SeaGreen3!30}-0.85 & \cellcolor{SeaGreen3!24}-1.23 & \cellcolor{SeaGreen3!25}-1.05 \\
    \midrule
    \cellcolor{SeaGreen3!77}\phantom{-}0.89 & \cellcolor{SeaGreen3!77}\phantom{-}0.85 & \cellcolor{SeaGreen3!80}\phantom{-}0.75 & \cellcolor{SeaGreen3!79}\phantom{-}0.74 \\
    \cellcolor{SeaGreen3!84}\phantom{-}1.13 & \cellcolor{SeaGreen3!73}\phantom{-}0.71 & \cellcolor{SeaGreen3!78}\phantom{-}0.66 & \cellcolor{SeaGreen3!73}\phantom{-}0.53 \\
    \cellcolor{SeaGreen3!90}\phantom{-}1.37 & \cellcolor{SeaGreen3!90}\phantom{-}1.34 & \cellcolor{SeaGreen3!90}\phantom{-}1.09 & \cellcolor{SeaGreen3!90}\phantom{-}1.11 \\
    \cellcolor{SeaGreen3!81}\phantom{-}1.05 & \cellcolor{SeaGreen3!78}\phantom{-}0.90 & \cellcolor{SeaGreen3!83}\phantom{-}0.86 & \cellcolor{SeaGreen3!85}\phantom{-}0.95 \\
    \midrule
    \cellcolor{SeaGreen3!74}\phantom{-}0.76 & \cellcolor{SeaGreen3!74}\phantom{-}0.74 & \cellcolor{SeaGreen3!71}\phantom{-}0.44 & \cellcolor{SeaGreen3!75}\phantom{-}0.60 \\
    \bottomrule
\end{tabular}

\caption{
Results (average quality estimation scores) for using different LLMs (rows) as part of MT-breaker.
Left side shows translation quality (lower is better) and right side shows average of z-normalized diversity measures.
Averaged across all languages.
}
\label{tab:04-other_llm}
\end{table*}
\subsection{Data Quality}

So far we have investigated only the method's capability of breaking the machine translation model.
We now confirm the usability of the method for generating meaningful datasets.
Due to the fact that there is no accepted way of intrinsically evaluating a dataset, we propose to examine several different statistics about the dataset.
Prior works either do not investigate the data quality \citep{pombal2025zero}, rely on guarantees of the data creation process, such as grammatical or human edits \citep{isabelle-etal-2017-challenge,manakhimova-etal-2023-linguistically}, or resort to manual human inspection \citep{amrhein-etal-2022-aces,armannsson-etal-2024-killing}.

For a generated set of examples, we measure the following variables:
\begin{itemize}
\item Diversity (in embedding space, pairwise chrF \citealp{popovic-2015-chrf},\footnote{\texttt{ref:1|case:mix|eff:y|nc:6|nw:2|sp:no|v:2.5.1}} unique topics, unique words), to make sure that the texts does not collapse into one subtopic.
\item Diversity (errors), to make sure that the error mode is not simply being repeated.
\item Source and word length, to make sure the difficulty is not being trivially induced by extremely long segments.
\item Word rarity, to analyze if the difficulty stems from simply using rare words.
\item Syntax complexity, to analyze if the difficulty is due to convoluted sentence structures.
\item Grammaticality, to make sure the text is well-formed and not broken.
\item Naturalness, to make sure the text sounds like it could have been written by a human.
\end{itemize}
\noindent
See \Cref{sec:prompts} for details on how some of these variables are measured using a prompted LLM.

\smallskip

The results in \Cref{tab:01-data_quality} show that methods without an explicit treatment for diversity (Zeroshot, Zeroshot min, MT-breaker seedless) indeed lack in the diversity, though this effect is most prevalent in Zeroshot.
On the other hand, the methods without diversity guardrails lead to texts of higher difficulty.
This tradeoff between diversity and difficulty is shown in \Cref{fig:01-difficulty_pareto}.
If only the difficulty is the concern and not the diversity, MT-breaker seedless is an apt choice.
If, however, one desires to stay close to some existing dataset in terms of diversity but needs more difficulty, MT-breaker seeded provides a balance between the two.

\begin{table}[t]
\small
\centering
\begin{tabular}{c@{\hspace{2mm}}l *{5}{>{\centering\arraybackslash\hspace*{-0.8mm}}p{6mm}}}
    \toprule
    & & \multicolumn{5}{c}{Target model} \\
    &&
    \rotatebox[origin=c]{90}{Translate} &
    \rotatebox[origin=c]{90}{Gemma3} &
    \rotatebox[origin=c]{90}{Gemini2.0} &
    \rotatebox[origin=c]{90}{Gemini2.5} &
    \rotatebox[origin=c]{90}{o4-mini} \\
    \cmidrule{3-7}
    \multirow{5}{*}{\rotatebox[origin=c]{90}{\parbox{1.8cm}{\centering\bf MT-break.\\(seedless)}}}
    & Translate  & \cellcolor{Firebrick3!90}45.07 & \cellcolor{Firebrick3!57}68.86 & \cellcolor{Firebrick3!60}70.68 & \cellcolor{Firebrick3!57}70.15 & \cellcolor{Firebrick3!77}77.95 \\
    & Gemma3     & \cellcolor{Firebrick3!60}60.35 & \cellcolor{Firebrick3!90}54.55 & \cellcolor{Firebrick3!65}68.76 & \cellcolor{Firebrick3!66}66.92 & \cellcolor{Firebrick3!83}76.70 \\
    & Gemini2.0  & \cellcolor{Firebrick3!64}58.12 & \cellcolor{Firebrick3!64}65.75 & \cellcolor{Firebrick3!90}58.94 & \cellcolor{Firebrick3!67}66.18 & \cellcolor{Firebrick3!84}76.42 \\
    & Gemini2.5  & \cellcolor{Firebrick3!68}56.42 & \cellcolor{Firebrick3!66}64.89 & \cellcolor{Firebrick3!68}67.29 & \cellcolor{Firebrick3!90}57.09 & \cellcolor{Firebrick3!90}75.18 \\
    & Multi      & \cellcolor{Firebrick3!74}53.30 & \cellcolor{Firebrick3!79}59.39 & \cellcolor{Firebrick3!79}63.09 & \cellcolor{Firebrick3!77}62.44 & \cellcolor{Firebrick3!84}76.54 \\
    \midrule
    \multirow{5}{*}{\rotatebox[origin=c]{90}{\parbox{1.8cm}{\centering\bf Zeroshot\\(min)}}}
    & Translate  & \cellcolor{Firebrick3!35}72.89 & \cellcolor{Firebrick3!37}77.62 & \cellcolor{Firebrick3!39}78.87 & \cellcolor{Firebrick3!32}80.24 & \cellcolor{Firebrick3!50}83.66 \\
    & Gemma3     & \cellcolor{Firebrick3!35}73.17 & \cellcolor{Firebrick3!43}74.91 & \cellcolor{Firebrick3!42}77.45 & \cellcolor{Firebrick3!37}78.53 & \cellcolor{Firebrick3!57}82.22 \\
    & Gemini2.0  & \cellcolor{Firebrick3!33}74.17 & \cellcolor{Firebrick3!38}77.33 & \cellcolor{Firebrick3!42}77.52 & \cellcolor{Firebrick3!31}80.78 & \cellcolor{Firebrick3!48}83.96 \\
    & Gemini2.5  & \cellcolor{Firebrick3!35}73.08 & \cellcolor{Firebrick3!41}75.71 & \cellcolor{Firebrick3!44}76.90 & \cellcolor{Firebrick3!40}77.21 & \cellcolor{Firebrick3!56}82.33 \\
    & Multi      & \cellcolor{Firebrick3!46}67.46 & \cellcolor{Firebrick3!55}69.57 & \cellcolor{Firebrick3!62}69.64 & \cellcolor{Firebrick3!52}72.51 & \cellcolor{Firebrick3!71}79.14 \\
    \midrule
    \multirow{5}{*}{\rotatebox[origin=c]{90}{\parbox{1.8cm}{\centering\bf MT-break.\\(seeded)}}}
    & Translate  & \cellcolor{Firebrick3!47}66.97 & \cellcolor{Firebrick3!28}81.66 & \cellcolor{Firebrick3!27}83.20 & \cellcolor{Firebrick3!28}82.16 & \cellcolor{Firebrick3!38}86.05 \\
    & Gemma3     & \cellcolor{Firebrick3!29}75.95 & \cellcolor{Firebrick3!39}76.67 & \cellcolor{Firebrick3!28}82.85 & \cellcolor{Firebrick3!28}82.19 & \cellcolor{Firebrick3!40}85.66 \\
    & Gemini2.0  & \cellcolor{Firebrick3!31}74.90 & \cellcolor{Firebrick3!31}80.13 & \cellcolor{Firebrick3!43}77.02 & \cellcolor{Firebrick3!30}81.14 & \cellcolor{Firebrick3!44}84.77 \\
    & Gemini2.5  & \cellcolor{Firebrick3!31}74.83 & \cellcolor{Firebrick3!28}81.49 & \cellcolor{Firebrick3!27}83.35 & \cellcolor{Firebrick3!40}77.35 & \cellcolor{Firebrick3!38}86.17 \\
    & Multi      & \cellcolor{Firebrick3!40}70.38 & \cellcolor{Firebrick3!37}77.49 & \cellcolor{Firebrick3!40}78.29 & \cellcolor{Firebrick3!38}78.04 & \cellcolor{Firebrick3!42}85.19 \\
    \midrule
    & \bf Seeds & \cellcolor{Firebrick3!10}85.63 & \cellcolor{Firebrick3!10}89.26 & \cellcolor{Firebrick3!10}89.94 & \cellcolor{Firebrick3!10}89.23 & \cellcolor{Firebrick3!10}92.00 \\
    \bottomrule
\end{tabular}

\vspace{-2mm}
\caption{Results (average quality estimation scores) for difficulty transfer between target machine translation models. MT-breaker and Zeroshot are optimized towards one MT (rows) but the obtained examples are evaluated on another MT (columns).
Averaged across all languages.
}
\label{tab:02-transfer_model}

\vspace{-3mm}
\end{table}

\begin{table}[t]
\small
\centering
\begin{tabular}{c@{\hspace{2mm}}l *{5}{>{\centering\arraybackslash}p{7mm}}}
    \toprule
    & & \multicolumn{5}{c}{Language Pair} \\
    && EnCs & EnDe & EnZh & EnVi & EnPl \\
    \midrule
    \multirow{5}{*}{\rotatebox[origin=c]{90}{\parbox{1.8cm}{\centering\bf MT-breaker\\(seedless)}}}
    & EnCs & \cellcolor{Firebrick3!90}{42.28} & \cellcolor{Firebrick3!54}{74.40} & \cellcolor{Firebrick3!48}{76.23} & \cellcolor{Firebrick3!56}{71.50} & \cellcolor{Firebrick3!65}{60.61} \\
    & EnDe & \cellcolor{Firebrick3!74}{50.57} & \cellcolor{Firebrick3!90}{59.41} & \cellcolor{Firebrick3!72}{67.32} & \cellcolor{Firebrick3!76}{63.72} & \cellcolor{Firebrick3!82}{52.77} \\
    & EnZh & \cellcolor{Firebrick3!65}{55.63} & \cellcolor{Firebrick3!61}{71.49} & \cellcolor{Firebrick3!90}{60.68} & \cellcolor{Firebrick3!69}{66.32} & \cellcolor{Firebrick3!69}{58.65} \\
    & EnVi & \cellcolor{Firebrick3!63}{56.40} & \cellcolor{Firebrick3!58}{72.74} & \cellcolor{Firebrick3!61}{71.19} & \cellcolor{Firebrick3!90}{57.87} & \cellcolor{Firebrick3!71}{58.03} \\
    & EnPl & \cellcolor{Firebrick3!61}{57.22} & \cellcolor{Firebrick3!53}{74.66} & \cellcolor{Firebrick3!56}{73.04} & \cellcolor{Firebrick3!61}{69.66} & \cellcolor{Firebrick3!90}{49.34} \\
    \midrule
    \multirow{5}{*}{\rotatebox[origin=c]{90}{\parbox{1.8cm}{\centering\bf Zeroshot\\(min)}}}
    & EnCs & \cellcolor{Firebrick3!50}{63.42} & \cellcolor{Firebrick3!32}{83.54} & \cellcolor{Firebrick3!29}{83.02} & \cellcolor{Firebrick3!39}{78.56} & \cellcolor{Firebrick3!42}{71.17} \\
    & EnDe & \cellcolor{Firebrick3!57}{59.76} & \cellcolor{Firebrick3!41}{79.78} & \cellcolor{Firebrick3!41}{78.54} & \cellcolor{Firebrick3!49}{74.40} & \cellcolor{Firebrick3!50}{67.29} \\
    & EnZh & \cellcolor{Firebrick3!45}{66.04} & \cellcolor{Firebrick3!31}{83.90} & \cellcolor{Firebrick3!39}{79.24} & \cellcolor{Firebrick3!38}{79.08} & \cellcolor{Firebrick3!40}{71.93} \\
    & EnVi & \cellcolor{Firebrick3!48}{64.04} & \cellcolor{Firebrick3!29}{84.61} & \cellcolor{Firebrick3!33}{81.43} & \cellcolor{Firebrick3!39}{78.58} & \cellcolor{Firebrick3!44}{70.14} \\
    & EnPl & \cellcolor{Firebrick3!32}{72.91} & \cellcolor{Firebrick3!19}{88.62} & \cellcolor{Firebrick3!20}{86.28} & \cellcolor{Firebrick3!24}{84.60} & \cellcolor{Firebrick3!29}{77.13} \\
    \midrule
    \multirow{5}{*}{\rotatebox[origin=c]{90}{\parbox{1.8cm}{\centering\bf MT-breaker\\(seeded)}}}
    & EnCs & \cellcolor{Firebrick3!43}{66.95} & \cellcolor{Firebrick3!27}{85.42} & \cellcolor{Firebrick3!26}{84.28} & \cellcolor{Firebrick3!29}{82.68} & \cellcolor{Firebrick3!30}{76.60} \\
    & EnDe & \cellcolor{Firebrick3!30}{73.72} & \cellcolor{Firebrick3!40}{79.90} & \cellcolor{Firebrick3!25}{84.56} & \cellcolor{Firebrick3!27}{83.61} & \cellcolor{Firebrick3!27}{77.98} \\
    & EnZh & \cellcolor{Firebrick3!32}{72.74} & \cellcolor{Firebrick3!28}{85.15} & \cellcolor{Firebrick3!44}{77.56} & \cellcolor{Firebrick3!30}{82.23} & \cellcolor{Firebrick3!29}{77.04} \\
    & EnVi & \cellcolor{Firebrick3!33}{72.03} & \cellcolor{Firebrick3!28}{85.01} & \cellcolor{Firebrick3!29}{83.12} & \cellcolor{Firebrick3!43}{77.04} & \cellcolor{Firebrick3!31}{76.04} \\
    & EnPl & \cellcolor{Firebrick3!33}{72.39} & \cellcolor{Firebrick3!30}{84.18} & \cellcolor{Firebrick3!26}{84.18} & \cellcolor{Firebrick3!28}{83.11} & \cellcolor{Firebrick3!42}{71.06} \\
    \midrule
    \multicolumn{2}{l}{\bf Seeds} & \cellcolor{Firebrick3!10}{84.19} & \cellcolor{Firebrick3!10}{92.50} & \cellcolor{Firebrick3!10}{90.01} & \cellcolor{Firebrick3!10}{90.31} & \cellcolor{Firebrick3!10}{85.56} \\
    \bottomrule
\end{tabular}

\caption{
Results (average quality estimation scores) for difficulty transfer between language pair directions. MT-breaker is optimized on one language pair (rows) but the obtained examples are evaluated on another language pair (columns). Highlights are normalized within columns.
Averaged across all models.
}
\label{tab:02-transfer_language}

\vspace{-3mm}
\end{table}

\subsection{Transfer Between MT Models}
\label{sec:experiments_transfer_model}

When MT-breaker targets a specific model, that model's performance is most significantly impacted, indicated by the lowest scores on the diagonal in \Cref{tab:02-transfer_model}, especially in contrast to the low difficulty Seeds.
While the difficulty does transfer to other models, the difficulty on other models is generally on par with or slightly lower than Zeroshot.
For this specific experiment only due to compute constraints, we also include \href{https://platform.openai.com/docs/models/o4-mini#model=gpt-4o-mini-2024-07-18}{o4-mini} as an additional machine translation model on which we evaluate the generated texts.
The model's performance has similar patterns to the other models, with most difficult examples being generated by MT-breaker (seedless).

Targeting the weakest of the models, Google Translate, generally leads to lower difficulty for other models, though this effect is not consistent when targeting more performant models.

The Seeds baseline shows similar performance across Gemma 3, Gemini 2.0 and Gemini 2.5 (89.3, 89.9, 89.2).
However, the Zeroshot approach reveals greater differences (74.9, 77.5, 77.2), and MT-breaker accentuates these distinctions even further (54.6, 58.9, 57.1).
This shows that difficult testsets or evaluation setups are useful in better benchmarking of models because larger gaps between models generally lead to more statistically significant outcomes.

Finally, we also include a Multi version of MT-breaker, that is optimized against Google Translate, Gemma 3, Gemini 2.0 and Gemini 2.5 at the same time.
This modification simply includes the outputs of all models instead of just one and for the final selection takes the average score across all models.
This approach does not always lead to the highest difficulty for the individual models, but it is at worst second best and notably better than when transferring from other individual models.
It also leads to higher difficulty for o4-mini, a machine translation model that was not part of the optimization, which suggests that the Multi version of MT-breaker taps into a more general notion of difficulty rather than model-specific one.

\subsection{Transfer between LLMs}
\label{sec:experiments_transfer_llms}

We now explore how the choice of the LLM in $\mathrm{LLM}_\mathrm{step}$ in \Cref{alg:mt_breaker} influences the generated data.
This is important, as LLMs can have blind spots, such as not being able to generate difficult texts for themselves.
For this, we generated the difficult texts using various LLMs that also ultimately serve as the machine translation models.
The results in \Cref{tab:04-other_llm} suggest that there is no discernible bias.
However, it shows that the outputs are systematically different.
For example, when optimized with Gemini 2.0 (Zeroshot min, MT-breaker seeded, MT-breaker seedless), the texts seem to be more difficult.
However, this comes at the expense of much lower diversity, which supports the tradeoff findings of \Cref{sec:experiments_difficulty}.

\subsection{Transfer Between Languages}
\label{sec:experiments_transfer_language}

The true translation difficulty is certainly a function of both the source and target languages, because some concepts might be more easily expressed in some languages than others.
However, we pose that it is mostly the property of the source language.
When generating difficult texts for a specific target language with MT-breaker, the resulting examples are most challenging for that particular language.
As shown in \Cref{tab:02-transfer_language}, the lowest translation quality scores consistently appear on the diagonal, where the optimization and evaluation languages match.
While some difficulty transfers to other languages, the effect is most pronounced on the targeted one.
In contrast, the Zeroshot (min) method shows no discernible or consistent pattern when transferring across different languages, despite the language direction being part of the prompt.
Still, even in the non-target languages, the MT-breaker (seedless) leads to higher difficulty than Zeroshot (min).

Translations from other languages into English show similar ranking of methods, as shown in Appendix \Cref{tab:03-xxen}.

\begin{table}[ht]
\small
\centering
\begin{tabular}{c@{\hspace{1mm}}lcccc}
\toprule
\bf QE
& \bf Translator
& \rotatebox{90}{\textbf{Seeds}}
& \rotatebox{90}{\makecell[l]{\bf Zeroshot \\ (min)}}
& \rotatebox{90}{\makecell[l]{\bf MT-breaker \\ (seedless)}}
& \rotatebox{90}{\makecell[l]{\bf MT-breaker \\ (seeded)}} \\
\midrule
\parbox[t]{2mm}{\multirow{3}{*}{\rotatebox[origin=c]{90}{Gem.\hspace{-3mm}}}}
& Gemini2.5
& \cellcolor{Firebrick3!10} 91.76 & \cellcolor{Firebrick3!32} 79.35 & \cellcolor{Firebrick3!90} 46.93 & \cellcolor{Firebrick3!42} 73.94 \\
& Human
& \cellcolor{Firebrick3!18} 87.22 & \cellcolor{Firebrick3!32} 79.28 & \cellcolor{Firebrick3!67} 59.63 & \cellcolor{Firebrick3!35} 77.66 \\\null\\[-0.1em]

\parbox[t]{2mm}{\multirow{2}{*}{\rotatebox[origin=c]{90}{MetricX\hspace{-2mm}}}}
& Gemini2.5
& \cellcolor{Firebrick3!10} 87.80 & \cellcolor{Firebrick3!50} 72.54 & \cellcolor{Firebrick3!90} 57.35 & \cellcolor{Firebrick3!36} 78.05 \\
& Human
& \cellcolor{Firebrick3!12} 86.88 & \cellcolor{Firebrick3!47} 73.60 & \cellcolor{Firebrick3!79} 61.60 & \cellcolor{Firebrick3!32} 79.50 \\\null\\[-0.1em]

\parbox[t]{2mm}{\multirow{2}{*}{\rotatebox[origin=c]{90}{MQM}}}
& Gemini2.5
& \cellcolor{Firebrick3!10} 93.80 & \cellcolor{Firebrick3!40} 89.08 & \cellcolor{Firebrick3!90} 81.08 & \cellcolor{Firebrick3!54} 86.84 \\
& Human
& \cellcolor{Firebrick3!16} 92.80 & \cellcolor{Firebrick3!19} 92.32 & \cellcolor{Firebrick3!39} 89.20 & \cellcolor{Firebrick3!45} 88.16 \\\null\\[-0.1em]

\parbox[t]{2mm}{\multirow{2}{*}{\rotatebox[origin=c]{90}{ESA}}}
& Gemini2.5
& \cellcolor{Firebrick3!18} 68.02 & \cellcolor{Firebrick3!34} 64.92 & \cellcolor{Firebrick3!88} 54.90 & \cellcolor{Firebrick3!32} 65.34 \\
& Human
& \cellcolor{Firebrick3!10} 69.38 & \cellcolor{Firebrick3!38} 64.18 & \cellcolor{Firebrick3!90} 54.54 & \cellcolor{Firebrick3!22} 67.17 \\
\bottomrule
\end{tabular}

\caption{Human translation of testsets and evaluation results. Shown is a subset where all humans provided the translation. See extended version in Appendix \Cref{tab:12-humeval_ext}.}
\label{tab:12-humeval}

\vspace{-3mm}
\end{table}

\subsection{Human Translations and Annotations}

To confirm the difficulty of generated texts and to determine if the texts difficult for machine translation models are also difficult for human translators, we take the 100 texts optimized against Gemini 2.5 in English$\rightarrow$Czech and English$\rightarrow$German and translate them with standard professional human translators.
Then, both the Gemini 2.5 and human translations are assessed with professional annotators for translation quality using a mixture of MQM \citep{freitag-etal-2021-experts} and ESA \citep{kocmi-etal-2024-error} protocols.\footnote{
Each translation gets a final score on the scale of 0 to 100, similar to ESA, but uses the full error taxonomy of MQM.
Annotators are translators from a professional language resource company (anonymized) with MQM Athena annotation interface.
}

The results in \Cref{tab:12-humeval} show similar difficulty patterns to \Cref{tab:01-data_quality}: MT-breaker (seedless) is the most difficult, followed by Zeroshot and MT-breaker seeded.
However, the annotators marked some segments as ``untranslateable'' for Zeroshot and MT-breaker (seedless) but less so for Seeds or MT-breaker (seeded).
Therefore, while ungrounded generations might lead to the highest difficulty, the examples might be too contrived for some applications.
Finally, there is very little difference between automatic and human translations, suggesting that the difficulty also transfers to human translators.
We include the error category breakdown in Appendix \Cref{tab:50-mqm_breakdown} which shows that \textit{Mistranslation} and \textit{Creative reinpretation} are much more common with MT-breaker and Zeroshot than Seeds.

\subsection{Interview with Experts}
\label{sec:human_qualitative}

To inform the design of MT-breaker, which aims to mimic how human experts find weaknesses in machine translation models, we conducted a qualitative study with seven MT practitioners with at least 2 years of academic or industry experience in developing or evaluating machine translation models.
The interview process involved two stages.

First, the participants were asked to interact with a machine translation model and to find its weaknesses.
This was done with a concurrent think-aloud protocol \citep{lewis1993task}, where the participants were commenting on their decisions and actions.
The interviews were not recorded.

\begin{enumerate}
    \item When you want to find weakpoints in a machine translation model, what is the first input you use?
    \item What kind of language you consider for input?
    \item Is your input choice influenced by the kinds of errors you're looking for apriori?
    \item Are your next inputs a modification of/related to the first one?
    \item How do you decide when to stop pursuing a specific direction before trying a new one?
\end{enumerate}

\paragraph{Findings.}
Based on qualitative interviews with machine translation experts, a common strategy for identifying model weaknesses involves an iterative and targeted approach.
Experts often begin with known challenging inputs, such as idioms and colloquialisms (I1, I5, I7), archaic expressions (I4), or specific domains like song lyrics (I3), drawing on their prior knowledge of typical MT failure points.
All interviewees (I1 to I7) at some point searched and content from the Internet as the input, which they then manually edited.
Upon finding a minor error, they tend to modify the input iteratively to provoke a larger failure, rather than switching to a new type of input immediately (I1 to I6).
The process is guided by their intuition about which part of the current input triggered an error.
Finally, some experts consider even non-standard input in order to trigger some unexpected behavior, like hallucinations.
This included ungrammatical sentences (I5, I6), wrong formatting (I3 to I7), repeating words multiple times (I5 to I7), mixed or incorrect language input (I2, I5, I6, I7), or random sequence of characters (I5, I6).
The patience in how long a certain input direction is being exploited when no error is found varies from 1-2 (I1, I5), 5 (I4), to 10 (I2, I3, I6).

Based on this, we designed MT-breaker to iteratively modify and exploit a specific input up to the depth of 10.
Contrary to some of the practices of the experts, we instructed LLM\textsubscript{step} to only make grammatical edits, so that we could fairly compare the methods.
Additionally, the quality estimation metrics are trained on mostly grammatical inputs, so they might not be able to accurately assess outputs with out-of-distribution error modes.

\section{Related Work}
\label{sec:related_work}

Our work builds upon three lines of research for creating challenging test sets for machine translation: selecting examples from existing corpora, generating examples from scratch, and creating adversarial examples.

\paragraph{Selecting examples.}
A common approach to building challenge sets is to select difficult examples from large, existing corpora.
\citet{chen-etal-2023-multifaceted} select sentences based on lexical rarity, structural complexity (sentence length), syntactic rarity (low-frequency parse trees), and model-based difficulty (high translation entropy).
\citet{proietti2025estimatingmachinetranslationdifficulty} train a source-based metric that predicts difficulty.
While effective, these selection-based methods are limited by the diversity of the source corpora and can fail to uncover novel or complex failure modes not present in naturally occurring text.
Lastly, \citet{zouhar2025select} show that samples with more errors are more useful for comparing models.

\paragraph{Generating examples.}
To overcome the constraints of selection, generative approaches create new examples.
Early and highly-controlled methods rely on manual, linguistically-informed construction \citep{isabelle-etal-2017-challenge}, or on semi-automated rule-based models to evaluate specific linguistic phenomena, which are then verified by human experts \citep{manakhimova-etal-2023-linguistically}.

Closest to our work is zero-shot benchmarking \citep{pombal2025zero}, which uses LLMs at scale to generate examples from scratch in a zero-shot manner but with a control of the domain and, importantly, difficulty.
However, such from-scratch generation often lacks sufficient difficulty and diversity for state-of-the-art models.
Our work combines the scalability of LLM generation with preserving diversity through seed texts and the targeted difficulty through an iterative feedback loop.

\paragraph{Adversarial examples.}
The goal of adversarial examples is not to create inherently difficult text, but to induce model failure through minimal, often imperceptible, perturbations to an input that is otherwise easy to translate.
\citet{zou-etal-2020-reinforced} use reinforcement learning to apply token-level edits that degrade translation quality while preserving the meaning of the source text.
Similarly, \citet{zhang-etal-2021-crafting} apply a sequence of heuristic edits to probe model brittleness to superficial input changes.

\paragraph{Interactive discovery.}
Several works have investigated mimicking the human expert who probes the model for weaknesses.
\citet{lu2025automated} propose Automated Capability Discovery, where one foundation model acts as a ``scientist'' to autonomously generate open-ended tasks (from a template) that reveal the capabilities and failures of another model.
Similarly, \citet{tjuatja2025behaviorbox} introduce BehaviorBox, a method to automatically discover fine-grained, interpretable features where one model systematically outperforms another.
None of these works, however, made use off the specifics of machine translation, where we have access to accurate quality estimation metrics.

\section{Conclusion}
\vspace{-1mm}

\paragraph{Discussion.}
The MT-breaker (intentionally) introduces a bias against the particular machine translation model which it is trying to break and find and amplify the weaknesses of.
Consequently, the generated dataset is tailored to that model's particular vulnerabilities and cannot be used for a fair comparison across different models.
Using this dataset to evaluate other models would be misleading, as it would disproportionately favor models that do not share the same specific weaknesses as the target model.
Instead, we suggest using MT-breaker in two of the following ways:
(1) Use the MT-breaker-generated dataset against a particular machine translation model $m_1$ to find its weaknesses, which can be used to hillclimb on.
(2) Compare the scores of two MT-breaker-generated datasets against $m_1$ and $m_2$ (i.e. try to break $m_1$ and $m_2$ independently) to discover the models' worst-case performances.

\paragraph{Summary.}
We introduced MT-breaker, a novel simple method that mimics human experts to iteratively generate texts that are difficult to translate.
This approach increases translation difficulty for specific machine translation models (confirmed by an expert human evaluation) while maintaining the naturalness and diversity found in the original seed texts.
While the difficulty is most pronounced for the targeted model and language, it also transfers to other models and languages.

\paragraph{Future work.}
MT-breaker uses signal from the specific model to obtain breaking inputs.
However, we hypothesize that some texts generated by both Zeroshot and MT-breaker methods may be difficult for humans to translate, not for machine translation models.
This is because many sources in the training data (Internet) discuss difficulty with respect to a human translator, so LLM\textsubscript{step} might mimic this.

\section*{Limitations}

We have relied heavily on the quality estimation metric as part of both the optimization objective and the quantity of interest.
To this end we used Zeroshot (min) variant, which uses a similar selection process to make the comparison more fair, and also confirm difficulty with human evaluation.

The Zeroshot method introduced by \citet{pombal2025zero} is conditioned on specific domains.
This could lead to more diverse outputs, though we did not include this in our implementation of Zeroshot so that it is more comparable to the other methods.

Lastly, the created texts fall outside of the expected user distribution (see Naturalness in \Cref{tab:01-data_quality}).
However, this is intentional as most texts are not challenging \citep{proietti2025estimatingmachinetranslationdifficulty} and we wish to focus on the more tail of the distribution with less likely but more difficult texts that reveal more the models' worst-case performances.

\section*{Ethics Statement}

The data used within the human study are sourced from the WMT 2024 testset \citep{kocmi-etal-2024-findings} and have thus already been screened for disturbing content.
The translators and annotators were fairly compensated, consented to participate in a research study, and no personal data was collected.

\bibliography{bibliography,anthology.min.bib}
\clearpage

\appendix

\begin{figure}[t]
\small
\centering
\begin{tabular}{@{\hspace{0.5mm}}l@{\hspace{2.5mm}}*{5}{>{\centering\arraybackslash\hspace*{-1.5mm}}p{5mm}}}
\toprule
& CsEn & DeEn & ZhEn & ViEn & PlEn \\
\midrule
MT-breaker (seedless) 
& \cellcolor{Firebrick3!90} 55.17 
& \cellcolor{Firebrick3!90} 63.78 
& \cellcolor{Firebrick3!90} 50.19 
& \cellcolor{Firebrick3!90} 63.58 
& \cellcolor{Firebrick3!90} 53.84 \\
Zeroshot (min) 
& \cellcolor{Firebrick3!75} 61.52 
& \cellcolor{Firebrick3!61} 74.30 
& \cellcolor{Firebrick3!58} 67.88 
& \cellcolor{Firebrick3!56} 76.05 
& \cellcolor{Firebrick3!66} 64.56 \\
MT-breaker\hspace{0.3mm}(seeded) 
& \cellcolor{Firebrick3!45} 74.85 
& \cellcolor{Firebrick3!44} 80.29 
& \cellcolor{Firebrick3!31} 82.97 
& \cellcolor{Firebrick3!48} 78.92 
& \cellcolor{Firebrick3!39} 76.15 \\
Seeds 
& \cellcolor{Firebrick3!10} 90.02 
& \cellcolor{Firebrick3!10} 92.81 
& \cellcolor{Firebrick3!10} 94.47 
& \cellcolor{Firebrick3!10} 92.75 
& \cellcolor{Firebrick3!10} 88.91 \\
\bottomrule
\end{tabular}

\caption{
Results for difficulty for translation into English, compare to diagonal in \Cref{tab:02-transfer_language}.
}
\label{tab:03-xxen}
\end{figure}

\section{Source-only MetricX}
\label{sec:srcqe_metricx}

We also trained a source-only version of MetricX to estimate the difficulty without the translation model's output.
Similar to Sentinel-src \citep{perrella-etal-2024-guardians,proietti2025estimatingmachinetranslationdifficulty} or PreCOMET \citep{zouhar2025select}, the input to the source-only MetricX is only the source text.
The model is trained to predict the average human assessment score aggregated across all translations for the given source in the data.
For example, if two models translate the same source and their human assessments are 76 and 98 MQM scores, the source-only model will be trained to produce 87.
The rest of the training follows the procedure of MetricX-23 \citep{juraska-etal-2023-metricx}.

\section{Prompts}
\label{sec:prompts}

\newcommand{\promptexample}[2]{
\noindent \textbf{#1.}
\setstretch{0.9}
{\fontsize{8.15pt}{8pt}\selectfont\tt #2 }
\medskip
}

The first prompt is the same across all methods:\smallskip
\promptexample{Zeroshot, MT-breaker}{
We want to find a text in \{LANG1\} that's exceptionally difficult
for a machine translation model to translate into \{LANG2\}.
The goal is to expose a wide range of translation errors and severely
challenge the MT model's capabilities.
The text should be approximately \{SEED\_LENGTH\} words.
At the end of your response write SOURCE |||<SOURCE\_TEXT>|||`.}

\promptexample{MT-breaker (qe), follow-up}{
This is the machine translation of the source text.\\
TRANSLATION |||\{TARGET\_TEXT\}|||\\
And this is the score of the translation: SCORE |||\{QE\_SCORE:.1f\}\%||| (0=bad translation, 100=perfect)
Try to make it even more difficult to translate. Modify at most 75\% of the best SOURCE (one with lowest score) so far.
}

\promptexample{Source text analysis}{
Analyze the following text and return the answer in JSON. We want to determine the following attributes:\\
- grammaticality: on a scale from 0 (complete gibbersish) to 100 (perfect, no errors).\\
- naturalness: on a scale from 0 (wholy unnatural) to 100 (fully human-like and would occur in a corpus).\\
- word rarity: on average, how rare are the words from 0 (average modern human would use this word daily) to 100 (average modern human would not understand the word).\\
- syntax complexity: on a scale from 0 (simplest possible sentence) to 100 (most complex and hard to understand).\\
- topics: list of 1 to 5 topics that the text is about.\\
Provide only the output in JSON and nothing else. The output should look like this (no extra backticks or newlines):\\
\{ \\
    "grammaticality": 90, \\
    "naturalness": 80, \\
    "word rarity": 50, \\
    "syntax complexity": 70, \\
    "topics": ["science", "technology"], \\
\}\\
The sentence to analyze is: \\
> \{SOURCE\_TEXT\}
}

\promptexample{Target text analysis}{
Analyze the following translation and return a list of reasons the translation might be incorrect (in JSON).
The options are: 
"idioms", "metaphor", "gender", "style", "ambiguity", "named entities",
"numerical expressions", "dates and times", "cultural nuances", "tone",
"context", "syntax", "semantics", "lexical choice", "register",
"collocations", "terminology", "cohesion and coherence",
"omissions", "additions", "misinterpretations", "negation", "tense and aspect",
"modality", "pronoun resolution", "punctuation", "formatting". \\
Feel free to add extra if you think it's appropriate. Output a list of reasons (even 0, depending on if the translation is erroneous).
Provide only the output in JSON and nothing else. The output should look like this (no extra backticks or newlines):\\
{\\
    "error\_modes": ["idiom", "gender", "style"], \\
} \\
The sentence to analyze is: \\
> \{SOURCE\_TEXT\} \\
The translation is: \\
> \{TARGET\_TEXT\}
}

\promptexample{LLM-as-QE}{
Evaluate the quality of the translation on a scale from 0 100. Roughly: \\
100 - Perfect \\
95 - Excellent (closely aligned with the source) \\
80 - Very good (minor style choice) \\
60 - Fair (some inaccuracies or fluency errors) \\
40 - Poor (multiple inaccuracies or fluency errors) \\
0 - Inadequate (unrelated, completely wrong) \\
First, think about all the errors in the translation
and their severity (very briefly, max few words per error).
At the end, output a single line in the format like as follows: 
`SCORE |||70.8|||`
The last line is important because it will be matched with a regex, so make sure to use the |.\\\\
SOURCE: |||\{SOURCE\_TEXT\}||| \\
TRANSLATION: |||\{TARGET\_TEXT\}|||
}

\clearpage

\begin{table*}[t]
\section{Data Examples}
\bigskip

\renewcommand{\arraystretch}{1.5}
\fontsize{8}{8}\selectfont
\centering
\begin{tabular}{ p{4.8cm}p{4.8cm}p{4.8cm} } \toprule
\bf\small Zeroshot (min) & \bf\small Seeds & \bf\small MTbreaker (seeded)\\
It is what it is. & Another one has been found! & Another rabbit hole of lies was found!\\
The grizzled grackle, having grabbled a grub, gargled a guttural greeting. Its gambrel legs, like spindly stilts, straddled a gnarled branch, its gaze a glassy glint. The gloaming gathered, a gossamer shroud, as the grotesque gargoyle, its grimace a grotesque geometry, glowered from the gable. Below, a gaggle of geese honked a garrulous goodnight, their glossy plumage a ghostly gleam in the waning light. & We're just going to let that sit and rehydrate for a couple minutes. With ultralight cooking, we're not actually cooking on the stove. We're primarily boiling water so that we can rehydrate our food. It's also one of the reasons why we use instant grits. Less cooking time, less fuel, and less weight. Oh yeah. So this is about what we've got going on here. Perfect consistency. & We're gonna let this grub rehydrate a touch. Ultralight cooking's not really about cooking, per se. We mostly just boil water to rehydrate our grub. That's why we use instant grits, man. Less cook time, fuel, and weight. Gnarly, a perfect consistency. So this is what we've got going on right here. Far out. We're gonna grub on this for a minute. It's gonna be so bomb, bro. Fire, no cap. Let's get this bread. It's bussin'. This slaps.\\
The quick brown fox jumps over the lazy dog. & I quit FB, removed all Meta and Pinterest tracker from my website and killed my private WhatsApp account. & I've quit FB and killed the private WhatsApp account to get rid of Meta's and Pinterest's gross ad-tracking tools that I hated with a fiery passion.\\
The wily, woolgathering wordsmith, having just quaffed a flagon of kvass, began weaving a web of wonton words. He spun a yarn about a gaggle of geese, each given a gimcrack gewgaw, gossiping about a gimpy gnu. Their gabbles grew grandiose, morphing into a grandiloquent gallimaufry of gobbledygook. The gnu, a grizzled grammarian, grimaced, groaning at the grammatical grotesqueries. This ghastly gibberish, he groused, was a genuine gerrymander of language, a grotesque gargoyle of grammar. What a galling, garrulous goose-chase! I’m totally knackered. & The Department of Justice's overall approach to corporate crimes has come under intense scrutiny from advocates. While watchdogs have lauded some steps the department has taken during Merrick Garland's tenure as attorney general - like finally establishing a database on corporate crime - Biden's DOJ has also leaned heavily on leniency agreements that allow companies to defer or avoid prosecution, and encouraged companies to scapegoat individual employees so as to avoid a broader charge. Kenneth Polite Jr., former assistant attorney general for DOJ's Criminal Division, revised the division's corporate enforcement policy so as to limit prosecutions. & The Department of Justice's overall approach to corporate crimes has come under intense scrutiny from advocates. While watchdogs have lauded some steps the department has taken during Merrick Garland's tenure as attorney general - like finally establishing a database on corporate crime - Biden's DOJ has also leaned heavily on leniency agreements that allow companies to defer or avoid prosecution, and encouraged companies to scapegoat individual employees so as to avoid a broader charge. Kenneth Polite Jr., former assistant attorney general for DOJ's Criminal Division, revised the division's corporate enforcement policy so as to limit prosecutions.\\
The jabberwocky, with eyes of flame, came whiffling through the tulgey wood, and burbled as it came! `Twas brillig, and the slithy toves did gyre and gimble in the wabe; all mimsy were the borogoves, and the mome raths outgrabe. & As they drew closer, the little heat signatures turned from blobs to more distinct shapes. As they drew closer, one of them ran for something and grabbed it before rusing back to the small ridge they were on. The figures dropped down into firing positions. & As they drew closer, the little heat signatures turned from blobs to more distinct shapes. As they drew closer, one of them ran for something and grabbed it before rusing back to the small ridge they were on. The figures dropped down into firing positions.\\
He bade them farewell and then he bade them all to hell. & Going back up tomorrow and we're doing stalls and slow flight. & Going back up tomorrow; we're doing stalls and spins, and then some unusual attitudes.\\
It is what it is. & Heheh not one but three! & Heheh not one, two, but three!!!11!oneeleven\\
The cantankerous clockmaker, a man of mercurial moods and intricate mechanisms, grumbled as he tinkered with a recalcitrant timepiece. Its delicate gears, a labyrinth of brass and steel, refused to cooperate, their stubborn silence echoing his own frustration. He muttered a litany of obscure horological terms, his voice a gravelly counterpoint to the ticking of a dozen other clocks. Each tick was a tiny hammer blow against his patience, a reminder of time's relentless march. He yearned for a world without time, a place of blissful stasis where gears wouldn't jam and springs wouldn't break. & 0430 Itania time. Cohren hated waking up this early. Mainly because it was his responsibility to wake everyone else up in his company. Silently stepping out of his bunk, he put on his combat fatigues, throwing on his assault vest, and packed up his belongings from his locker, being careful with a particular small box. He shoulded his rifle and put his helmet on a small hook connected to his belt. It had all its fixing, respirator, goggles, everything needed. After checking his ammo pouches, he quietly walked out of the barracks, mud and ice squelching under his boots. & 0430 Itania time, not that Cohren cared for punctuality. The wee hours bit, stingier than a scrooge's handshake. His lot: rousing the grunts. He crept from his bunk, a ghost in combat gear, shrugging on a vest heavy with unspoken promises. From his locker, he snagged his kit, a particular small box handled with the reverence of a holy relic. Rifle kissed his shoulder, helmet dangled from his belt like a shrunken head, bristling with fixings – respirator, goggles, the whole shebang. Ammo? Check. He slunk out, the pre-dawn muck groaning under his boots, a symphony of squelch.\\
The grizzled grackle, having grabbled a grub, gargled a guttural garble, a gratingly grandiloquent guffaw that grated on the great, grey granite gravestones. It was a ghastly, ghoulish glee, a grim gambol against the gloomy gloaming, a grotesque gesture against the gathering gloom. Gravely, the groundskeeper grumbled, “Gadzooks! That garrulous grackle gives me grave grievances!” & Stealthily I made my way towards the back of the store, murmurs and talking heard beyond a metal grate which took a few yanks to tear off. Climbing into the vent was the safest and quietest option for me right now; climbing my way through, I’d pick up on conversations and talks between strangely dressed men. I noticed Al inside, chained. Figures he’d get caught eventually. & With gusto, I wriggled my way toward the back of the entrepot, susurrations and parleying heard beyond a wrought iron grille which took a few goes to heave off. Subterfuge by vent was the soundest and hushest option for me right now; scrabbling my way along, I'd cotton on to consultations and confabs between queerly-garbed men. I clocked Al inside, shackled. Figures he'd get his collar felt eventually. I had to spring him.\\
\bottomrule\end{tabular}

\caption{Example of generated sources in English by Zeroshot, Seeds, and MT-breaker (seeded). Extends \Cref{tab:10-examples_small} from the main text.}
\label{tab:10-examples_all}
\end{table*}

\begin{table*}[t]
\renewcommand{\arraystretch}{1.5}
\fontsize{8}{8}\selectfont
\centering
\begin{tabular}{ p{15.5cm}} \toprule
\textbf{Step 0} \quad QE (Gemini): 65\%,  QE (MetricX): 72\% \\
Source:      Going back up tomorrow and we're doing stalls and slow flight. \\
Translation: Zítra jdeme zase nahoru a budeme dělat pádové rychlosti a pomalý let. \\[-0.2em]\null
\textbf{Step 1} \quad QE (Gemini): 45\%,  QE (MetricX): 79\% \\
Source:      Going back up tomorrow; we're doing stalls and spins, and then slow flight. \\
Translation: Zítra jdeme zase nahoru; budeme dělat pádové rychlosti a výkruty, a pak pomalý let. \\[-0.2em]\null
\textbf{Step 2} \quad QE (Gemini): 45\%,  QE (MetricX): 78\% \\
Source:      Going back up tomorrow; we're doing stalls and spins, and then some slow flight. \\
Translation: Zítra jdeme zase nahoru; budeme dělat pádové rychlosti a výkruty, a pak pomalý let. \\[-0.2em]\null
\textbf{Step 3} \quad QE (Gemini): 40\%,  QE (MetricX): 63\% {\color{red} \raisebox{0.4mm}{$\bullet$}} \\
Source:      Going back up tomorrow; we're doing stalls and spins, and then some unusual attitudes. \\
Translation: Zítra jdeme zase nahoru; budeme dělat \underline{pádové rychlosti} a \underline{výkruty}, a pak \underline{neobvyklé} polohy. \\[-0.2em]\null
\textbf{Step 4} \quad QE (Gemini): 65\%,  QE (MetricX): 74\% \\
Source:      Going back up tomorrow; we're doing stalls and spins, and then some unusual attitude recoveries. \\
Translation: Zítra jdeme zase nahoru; budeme dělat pádové a výkrutové manévry a pak nácvik neobvyklých poloh. \\[-0.2em]\null
\textbf{Step 5} \quad QE (Gemini): 45\%,  QE (MetricX): 66\% \\
Source:      Going back up tomorrow; we're doing stalls and spins, and then some unusual attitude recovery. \\
Translation: Zítra jdeme zase nahoru; budeme dělat pádové rychlosti a výkruty, a pak nácvik neobvyklých poloh. \\[-0.2em]\null
\textbf{Step 6} \quad QE (Gemini): 70\%,  QE (MetricX): 73\% \\
Source:      Going back up tomorrow; we're doing stalls and spins, and then some unusual attitude recovery maneuvers. \\
Translation: Zítra jdeme zase nahoru; budeme dělat pádové a výkrutové manévry a pak nácvik vybírání neobvyklých poloh. \\[-0.2em]\null
\textbf{Step 7} \quad QE (Gemini): 55\%,  QE (MetricX): 72\% \\
Source:      Going back up tomorrow; we're doing stalls and spins, and then some unusual attitude recovery maneuvers. \\
Translation: Zítra jdeme zase nahoru; budeme dělat pádové a výkrutové manévry a pak nácvik neobvyklých poloh. \\[-0.2em]\null
\textbf{Step 8} \quad QE (Gemini): 62\%,  QE (MetricX): 73\% \\
Source:      Going back up tomorrow; we're doing stalls and spins, and then some unusual attitude recovery maneuvers. \\
Translation: Zítra jdeme zase nahoru; budeme dělat pádové a výkrutové manévry a pak nácvik vybírání neobvyklých poloh. \\[-0.2em]\null
\textbf{Step 9} \quad QE (Gemini): 60\%,  QE (MetricX): 73\% \\
Source:      Going back up tomorrow; we're doing stalls and spins, and then some unusual attitude recovery maneuvers. \\
Translation: Zítra jdeme zase nahoru; budeme dělat pádové a výkrutové manévry a pak nácvik vybírání neobvyklých poloh. \\[-0.2em]\null
\textbf{Step 10} \quad QE (Gemini): 48\%,  QE (MetricX): 62\% \\
Source:      Going back up tomorrow; we're doing stalls and spins, and then some unusual attitude recovery maneuvers. \\
Translation: Zítra jdeme zase nahoru; budeme dělat pády a výkruty a pak nácvik neobvyklých poloh. \\
\bottomrule \end{tabular}

\caption{Example of one run of MT-breaker across 10 steps for English$\rightarrow$Czech. Step 3 {\color{red} \raisebox{0.3mm}{$\bullet$}} is selected as the final candidate because of lowest average score. \underline{Underlined} are major errors.}
\label{tab:10-examples_history_encs}
\end{table*}

\begin{table*}[t]
\renewcommand{\arraystretch}{1.5}
\fontsize{8}{8}\selectfont
\centering
\begin{tabular}{ p{15.5cm}} \toprule
\textbf{Step 0} \quad QE (Gemini): 100\%,  QE (MetricX): 88\% \\
Source:      Going back up tomorrow and we're doing stalls and slow flight. \\
Translation: Morgen geht es wieder hoch und wir machen Strömungsabrisse und Langsamflug. \\[-0.2em]\null
\textbf{Step 1} \quad QE (Gemini): 82\%,  QE (MetricX): 80\% \\
Source:      Going back up tomorrow; we're doing stalls and coffin corner. \\
Translation: Morgen geht es wieder hoch; wir machen Strömungsabrisse und den kritischen Geschwindigkeitsbereich. \\[-0.2em]\null
\textbf{Step 2} \quad QE (Gemini): 100\%,  QE (MetricX): 82\% \\
Source:      Going back up tomorrow; we're doing stalls and coffin corner practice. \\
Translation: Morgen geht es wieder hoch; wir machen Strömungsabriss- und Coffin Corner-Übungen. \\[-0.2em]\null
\textbf{Step 3} \quad QE (Gemini): 100\%,  QE (MetricX): 83\% \\
Source:      Going back up tomorrow; we're doing stalls and coffin corner practice drills. \\
Translation: Morgen geht es wieder hoch; wir machen Übungen zu Strömungsabrissen und zum "Coffin Corner". \\[-0.2em]\null
\textbf{Step 4} \quad QE (Gemini): 95\%,  QE (MetricX): 87\% \\
Source:      Going back up tomorrow; we're doing stalls and coffin corner practice drills to nail it. \\
Translation: Morgen geht es wieder hoch; wir machen Übungen zu Strömungsabrissen und dem kritischen Geschwindigkeitsbereich, um das hinzubekommen. \\[-0.2em]\null
\textbf{Step 5} \quad QE (Gemini): 85\%,  QE (MetricX): 74\% {\color{red} \raisebox{0.4mm}{$\bullet$}} \\
Source:      Going back up tomorrow; we're doing stalls and coffin corner practice drills to nail it on the checkride. \\
Translation: Morgen geht es wieder hoch; wir machen Übungen zu Strömungsabrissen und dem kritischen \underline{Geschwindigkeitsbereich}, um das bei der Prüfungs\underline{fahrt} perfekt hinzubekommen. \\[-0.2em]\null
\textbf{Step 6} \quad QE (Gemini): 95\%,  QE (MetricX): 73\% \\
Source:      Going back up tomorrow; we're doing stalls and coffin corner practice drills to nail it on the checkride exam. \\
Translation: Morgen geht es wieder hoch; wir machen Übungen zu Strömungsabrissen und dem kritischen Geschwindigkeitsbereich, um das bei der Prüfungsfahrt perfekt hinzubekommen. \\[-0.2em]\null
\textbf{Step 7} \quad QE (Gemini): 75\%,  QE (MetricX): 76\% \\
Source:      Going back up tomorrow; we're doing stalls and coffin corner practice drills to nail it on the checkride exam...wish me luck. \\
Translation: Morgen geht's wieder hoch; wir machen Übungen zu Strömungsabrissen und dem kritischen Geschwindigkeitsbereich, um das bei der Prüfungsfahrt hinzubekommen...drückt mir die Daumen. \\[-0.2em]\null
\textbf{Step 8} \quad QE (Gemini): 88\%,  QE (MetricX): 74\% \\
Source:      Going back up tomorrow; we're doing stalls and coffin corner practice drills to nail it on the checkride exam...wish me luck...not. \\
Translation: Morgen geht's wieder hoch; wir machen Übungen zu Strömungsabrissen und dem kritischen Geschwindigkeitsbereich, um das bei der Prüfungsfahrt hinzukriegen ... wünscht mir Glück ... nicht. \\[-0.2em]\null
\textbf{Step 9} \quad QE (Gemini): 70\%,  QE (MetricX): 71\% \\
Source:      Going back up tomorrow; we're doing stalls and coffin corner practice drills to nail it on the checkride exam...not. \\
Translation: Morgen geht's wieder hoch; wir machen Übungen zu Strömungsabrissen und dem Sargkurvenbereich, um das bei der Prüfungsfahrt perfekt hinzubekommen...nicht. \\[-0.2em]\null
\textbf{Step 10} \quad QE (Gemini): 72\%,  QE (MetricX): 73\% \\
Source:      Going back up tomorrow; we're doing stalls and coffin corner practice drills to nail it on the checkride exam...not! \\
Translation: Morgen geht's wieder hoch; wir machen Übungen zu Strömungsabrissen und dem kritischen Geschwindigkeitsbereich, um es bei der Prüfungsfahrt perfekt hinzubekommen...nicht! \\
\bottomrule \end{tabular}

\caption{Example of one run of MT-breaker across 10 steps for English$\rightarrow$German. Step 5 {\color{red} \raisebox{0.3mm}{$\bullet$}} is selected as the final candidate because of lowest average score. \underline{Underlined} are major errors.}
\label{tab:10-examples_history_ende}
\end{table*}

\clearpage

\begin{table*}
\section{Human Evaluation}
\vspace{3cm}

\small
\centering
\begin{tabular}{l@{\hspace{1mm}}lcccc}

\toprule
\multicolumn{2}{l}{EnCs+EnDe average} \\[-1em]
\bf QE 
& \bf Translator & \rotatebox{90}{\textbf{Seeds}}
& \rotatebox{90}{\makecell[l]{\bf Zeroshot \\ (min)}}
& \rotatebox{90}{\makecell[l]{\bf MT-breaker \\ (seedless)}}
& \rotatebox{90}{\makecell[l]{\bf MT-breaker \\ (seeded)}} \\
\midrule
Gemini & Gemini2.5
& \cellcolor{Firebrick3!10} 91.76 & \cellcolor{Firebrick3!32} 79.35 & \cellcolor{Firebrick3!90} 46.93 & \cellcolor{Firebrick3!42} 73.94 \\
Gemini & Human
& \cellcolor{Firebrick3!18} 87.22 & \cellcolor{Firebrick3!32} 79.28 & \cellcolor{Firebrick3!68} 59.36 & \cellcolor{Firebrick3!35} 77.66 \\\null\\[-0.1em]
MetricX & Gemini2.5
& \cellcolor{Firebrick3!10} 87.80 & \cellcolor{Firebrick3!50} 72.54 & \cellcolor{Firebrick3!90} 57.35 & \cellcolor{Firebrick3!36} 78.05 \\
MetricX & Human
& \cellcolor{Firebrick3!12} 86.88 & \cellcolor{Firebrick3!47} 73.60 & \cellcolor{Firebrick3!79} 61.60 & \cellcolor{Firebrick3!32} 79.50 \\\null\\[-0.1em]
MQM
& Gemini2.5
& \cellcolor{Firebrick3!10} 93.80 & \cellcolor{Firebrick3!40} 89.08 & \cellcolor{Firebrick3!90} 81.08 & \cellcolor{Firebrick3!54} 86.84 \\
& Human
& \cellcolor{Firebrick3!16} 92.80 & \cellcolor{Firebrick3!19} 92.32 & \cellcolor{Firebrick3!39} 89.20 & \cellcolor{Firebrick3!45} 88.16 \\\null\\[-0.1em]
ESA & Gemini2.5
& \cellcolor{Firebrick3!17} 68.02 & \cellcolor{Firebrick3!34} 64.92 & \cellcolor{Firebrick3!88} 54.90 & \cellcolor{Firebrick3!32} 65.34 \\
ESA & Human
& \cellcolor{Firebrick3!10} 69.38 & \cellcolor{Firebrick3!38} 64.18 & \cellcolor{Firebrick3!90} 54.54 & \cellcolor{Firebrick3!22} 67.17 \\\null\\[-0.1em]
Gemini & Gemini2.5 (NS)
& 92.00 & 40.25 & 57.79 & 55.67 \\
MetricX & Gemini2.5 (NS)
& 95.71 & 51.94 & 64.33 & 71.96 \\
NS & Human
& 0.5\% & 4.0\% & 28.5\% & 1.5\% \\
\bottomrule
\end{tabular}

\vspace{1em}

\begin{tabular}{l@{\hspace{1mm}}lcccc}
\toprule
\multicolumn{2}{l}{EnCs} \\[-1em]
\bf QE & \bf Translator & \rotatebox{90}{\textbf{Seeds}}
& \rotatebox{90}{\makecell[l]{\bf Zeroshot \\ (min)}}
& \rotatebox{90}{\makecell[l]{\bf MT-breaker \\ (seedless)}}
& \rotatebox{90}{\makecell[l]{\bf MT-breaker \\ (seeded)}} \\
\midrule
Gemini & Gemini2.5
& \cellcolor{Firebrick3!10} 91.78 & \cellcolor{Firebrick3!44} 76.66 & \cellcolor{Firebrick3!90} 56.64 & \cellcolor{Firebrick3!53} 73.11 \\
Gemini & Human
& \cellcolor{Firebrick3!24} 85.84 & \cellcolor{Firebrick3!46} 75.77 & \cellcolor{Firebrick3!68} 66.12 & \cellcolor{Firebrick3!49} 74.84 \\\null\\[-0.1em]
MetricX & Gemini2.5
& \cellcolor{Firebrick3!10} 83.11 & \cellcolor{Firebrick3!52} 64.22 & \cellcolor{Firebrick3!90} 47.38 & \cellcolor{Firebrick3!41} 69.42 \\
MetricX & Human
& \cellcolor{Firebrick3!15} 80.93 & \cellcolor{Firebrick3!56} 62.35 & \cellcolor{Firebrick3!78} 52.70 & \cellcolor{Firebrick3!39} 70.22 \\\null\\[-0.1em]
MQM & Gemini2.5
& \cellcolor{Firebrick3!14} 90.04 & \cellcolor{Firebrick3!34} 85.40 & \cellcolor{Firebrick3!90} 72.44 & \cellcolor{Firebrick3!64} 78.48 \\
MQM & Human
& \cellcolor{Firebrick3!15} 89.84 & \cellcolor{Firebrick3!10} 91.08 & \cellcolor{Firebrick3!11} 90.96 & \cellcolor{Firebrick3!49} 82.08 \\\null\\[-0.1em]
ESA & Gemini2.5
& \cellcolor{Firebrick3!34} 56.82 & \cellcolor{Firebrick3!30} 57.19 & \cellcolor{Firebrick3!90} 51.94 & \cellcolor{Firebrick3!46} 55.80 \\
ESA & Human
& \cellcolor{Firebrick3!10} 58.94 & \cellcolor{Firebrick3!27} 57.46 & \cellcolor{Firebrick3!41} 56.26 & \cellcolor{Firebrick3!30} 57.15 \\\null\\[-0.1em]
Gemini & Gemini2.5 (NS)
& - & 00.00 & 19.50 & - \\
MetricX & Gemini2.5 (NS)
& - & 10.87 & 20.42 & - \\
MQM & - & 80.00 & 40.00 & - \\
ESA & Gemini2.5 (NS)
& - & 52.00 & 49.00 & - \\
NS & Human
& 0.0\% & 3.0\% & 4.0\% & 0.0\% \\
\bottomrule
\end{tabular}
\begin{tabular}{lcccc}
\toprule
\multicolumn{2}{l}{EnDe} \\[-1em]
& \rotatebox{90}{\textbf{Seeds}}
& \rotatebox{90}{\makecell[l]{\bf Zeroshot \\ (min)}}
& \rotatebox{90}{\makecell[l]{\bf MT-breaker \\ (seedless)}}
& \rotatebox{90}{\makecell[l]{\bf MT-breaker \\ (seeded)}} \\
\midrule
& \cellcolor{Firebrick3!10} 91.74 & \cellcolor{Firebrick3!22} 82.10 & \cellcolor{Firebrick3!90} 27.11 & \cellcolor{Firebrick3!31} 74.79 \\
& \cellcolor{Firebrick3!14} 88.60 & \cellcolor{Firebrick3!21} 82.87 & \cellcolor{Firebrick3!67} 45.55 & \cellcolor{Firebrick3!24} 80.58 \\\null\\[-0.1em]
& \cellcolor{Firebrick3!12} 92.53 & \cellcolor{Firebrick3!72} 81.04 & \cellcolor{Firebrick3!90} 77.71 & \cellcolor{Firebrick3!41} 86.95 \\
& \cellcolor{Firebrick3!10} 92.89 & \cellcolor{Firebrick3!51} 85.09 & \cellcolor{Firebrick3!79} 79.79 & \cellcolor{Firebrick3!30} 89.07 \\\null\\[-0.1em]
& \cellcolor{Firebrick3!10} 97.60 & \cellcolor{Firebrick3!48} 92.76 & \cellcolor{Firebrick3!71} 89.82 & \cellcolor{Firebrick3!29} 95.20 \\
& \cellcolor{Firebrick3!25} 95.76 & \cellcolor{Firebrick3!42} 93.56 & \cellcolor{Firebrick3!90} 87.48 & \cellcolor{Firebrick3!36} 94.28 \\\null\\[-0.1em]
& \cellcolor{Firebrick3!12} 79.21 & \cellcolor{Firebrick3!32} 72.65 & \cellcolor{Firebrick3!79} 57.13 & \cellcolor{Firebrick3!25} 74.89 \\
& \cellcolor{Firebrick3!10} 79.82 & \cellcolor{Firebrick3!37} 70.90 & \cellcolor{Firebrick3!90} 53.53 & \cellcolor{Firebrick3!18} 77.19 \\\null\\[-0.1em]
& 92.00 & 64.40 & 60.68 & 55.67 \\
& 95.71 & 76.59 & 67.64 & 71.96 \\
& 100.00 & 96.80 & 85.16 & 94.68 \\
& 51.00 &  66.00 & 54.71 & 65.00 \\
& 1.0\% & 5.0\% & 53.0\% & 3.0\% \\
\bottomrule
\end{tabular}

\caption{Human translation and annotations. QE data is shown on a subset where all humans provided the translation and `NS' shows the segments that were not translated by human translators. Tables are split across two language directions (EnCs and EnDe) and also averaged. The MQM is converted to 100+4$\cdot$MQM to be on similar scale as MetricX. The tables xtend \Cref{tab:12-humeval}.}
\label{tab:12-humeval_ext}
\end{table*}

\begin{table*}
\small
\centering
\setlength{\tabcolsep}{2pt}
\begin{tabular}{ll >{\raggedleft\arraybackslash}p{11mm} c >{\raggedleft\arraybackslash}p{11mm}>{\raggedleft\arraybackslash}p{11mm} >{\raggedleft\arraybackslash}p{11mm}>{\raggedleft\arraybackslash}p{11mm}}
\toprule
& \bf Error type & All && EnCs & EnDe & Gemini2.5 & Human \\
\midrule
\parbox[t]{2mm}{\multirow{2}{*}{\rotatebox[origin=c]{90}{\bf Seeds}}} & Accuracy/Mistranslation & \cellcolor{Firebrick3!62} 18.5\% &  & \cellcolor{Firebrick3!73} 21.8\% & \cellcolor{Firebrick3!47} 14.1\% & \cellcolor{Firebrick3!62} 18.6\% & \cellcolor{Firebrick3!61} 18.3\% \\
 & Style/Unnatural or awkward & \cellcolor{Firebrick3!54} 16.1\% &  & \cellcolor{Firebrick3!74} 22.1\% & \cellcolor{Firebrick3!27} 8.1\% & \cellcolor{Firebrick3!56} 16.7\% & \cellcolor{Firebrick3!52} 15.5\% \\
 & Accuracy/Creative Reinterpretation & \cellcolor{Firebrick3!20} 6.1\% &  & \cellcolor{Firebrick3!22} 6.5\% & \cellcolor{Firebrick3!19} 5.6\% & \cellcolor{Firebrick3!13} 3.8\% & \cellcolor{Firebrick3!28} 8.3\% \\
 & Fluency/Grammar & \cellcolor{Firebrick3!14} 4.1\% &  & \cellcolor{Firebrick3!23} 6.8\% & \cellcolor{Firebrick3!1} 0.4\% & \cellcolor{Firebrick3!14} 4.2\% & \cellcolor{Firebrick3!13} 4.0\% \\
 & Fluency/Punctuation & \cellcolor{Firebrick3!11} 3.3\% &  & \cellcolor{Firebrick3!8} 2.3\% & \cellcolor{Firebrick3!16} 4.7\% & \cellcolor{Firebrick3!9} 2.7\% & \cellcolor{Firebrick3!13} 4.0\% \\
 & Style/Bad sentence structure & \cellcolor{Firebrick3!15} 4.4\% &  & \cellcolor{Firebrick3!20} 5.9\% & \cellcolor{Firebrick3!9} 2.6\% & \cellcolor{Firebrick3!19} 5.7\% & \cellcolor{Firebrick3!11} 3.2\% \\
 & Accuracy/Source language fragment & \cellcolor{Firebrick3!7} 2.0\% &  & \cellcolor{Firebrick3!2} 0.7\% & \cellcolor{Firebrick3!13} 3.8\% & \cellcolor{Firebrick3!9} 2.7\% & \cellcolor{Firebrick3!5} 1.4\% \\
 & Accuracy/Omission & \cellcolor{Firebrick3!9} 2.6\% &  & \cellcolor{Firebrick3!5} 1.6\% & \cellcolor{Firebrick3!13} 3.8\% & \cellcolor{Firebrick3!9} 2.7\% & \cellcolor{Firebrick3!8} 2.5\% \\
 & Fluency/Spelling & \cellcolor{Firebrick3!6} 1.8\% &  & \cellcolor{Firebrick3!7} 2.0\% & \cellcolor{Firebrick3!6} 1.7\% & \cellcolor{Firebrick3!0} 0.0\% & \cellcolor{Firebrick3!12} 3.6\% \\
 & Accuracy/Gender Mismatch & \cellcolor{Firebrick3!3} 0.9\% &  & \cellcolor{Firebrick3!5} 1.6\% & \cellcolor{Firebrick3!0} 0.0\% & \cellcolor{Firebrick3!1} 0.4\% & \cellcolor{Firebrick3!5} 1.4\% \\
 & Accuracy/Addition & \cellcolor{Firebrick3!1} 0.4\% &  & \cellcolor{Firebrick3!1} 0.3\% & \cellcolor{Firebrick3!1} 0.4\% & \cellcolor{Firebrick3!0} 0.0\% & \cellcolor{Firebrick3!2} 0.7\% \\
 & Style/Archaic or obscure word choice & \cellcolor{Firebrick3!2} 0.6\% &  & \cellcolor{Firebrick3!3} 1.0\% & \cellcolor{Firebrick3!0} 0.0\% & \cellcolor{Firebrick3!4} 1.1\% & \cellcolor{Firebrick3!0} 0.0\% \\
 & Terminology/Inappropriate for context & \cellcolor{Firebrick3!3} 0.9\% &  & \cellcolor{Firebrick3!2} 0.7\% & \cellcolor{Firebrick3!4} 1.3\% & \cellcolor{Firebrick3!3} 0.8\% & \cellcolor{Firebrick3!4} 1.1\% \\
\bottomrule
\end{tabular}
\smallskip

\begin{tabular}{ll >{\raggedleft\arraybackslash}p{11mm} c >{\raggedleft\arraybackslash}p{11mm}>{\raggedleft\arraybackslash}p{11mm} >{\raggedleft\arraybackslash}p{11mm}>{\raggedleft\arraybackslash}p{11mm}}
\toprule
& \bf Error type & All && EnCs & EnDe & Gemini2.5 & Human \\
\midrule
\parbox[t]{2mm}{\multirow{5}{*}{\rotatebox[origin=c]{90}{\bf Zeroshot (min)}}} & Accuracy/Mistranslation & \cellcolor{Firebrick3!93} 27.8\% &  & \cellcolor{Firebrick3!91} 27.2\% & \cellcolor{Firebrick3!95} 28.4\% & \cellcolor{Firebrick3!100} 31.1\% & \cellcolor{Firebrick3!80} 24.1\% \\
 & Style/Unnatural or awkward & \cellcolor{Firebrick3!49} 14.6\% &  & \cellcolor{Firebrick3!62} 18.8\% & \cellcolor{Firebrick3!32} 9.7\% & \cellcolor{Firebrick3!51} 15.4\% & \cellcolor{Firebrick3!46} 13.7\% \\
 & Accuracy/Creative Reinterpretation & \cellcolor{Firebrick3!34} 10.2\% &  & \cellcolor{Firebrick3!31} 9.2\% & \cellcolor{Firebrick3!38} 11.4\% & \cellcolor{Firebrick3!27} 8.2\% & \cellcolor{Firebrick3!41} 12.4\% \\
 & Fluency/Grammar & \cellcolor{Firebrick3!14} 4.3\% &  & \cellcolor{Firebrick3!23} 7.0\% & \cellcolor{Firebrick3!4} 1.3\% & \cellcolor{Firebrick3!16} 4.9\% & \cellcolor{Firebrick3!12} 3.7\% \\
 & Fluency/Punctuation & \cellcolor{Firebrick3!13} 3.9\% &  & \cellcolor{Firebrick3!15} 4.4\% & \cellcolor{Firebrick3!11} 3.4\% & \cellcolor{Firebrick3!16} 4.9\% & \cellcolor{Firebrick3!10} 2.9\% \\
 & Style/Bad sentence structure & \cellcolor{Firebrick3!7} 2.2\% &  & \cellcolor{Firebrick3!10} 2.9\% & \cellcolor{Firebrick3!4} 1.3\% & \cellcolor{Firebrick3!7} 2.2\% & \cellcolor{Firebrick3!7} 2.1\% \\
 & Accuracy/Source language fragment & \cellcolor{Firebrick3!11} 3.3\% &  & \cellcolor{Firebrick3!5} 1.5\% & \cellcolor{Firebrick3!18} 5.5\% & \cellcolor{Firebrick3!19} 5.6\% & \cellcolor{Firebrick3!3} 0.8\% \\
 & Accuracy/Omission & \cellcolor{Firebrick3!5} 1.4\% &  & \cellcolor{Firebrick3!7} 2.2\% & \cellcolor{Firebrick3!1} 0.4\% & \cellcolor{Firebrick3!0} 0.0\% & \cellcolor{Firebrick3!10} 2.9\% \\
 & Fluency/Spelling & \cellcolor{Firebrick3!7} 2.2\% &  & \cellcolor{Firebrick3!7} 2.2\% & \cellcolor{Firebrick3!7} 2.1\% & \cellcolor{Firebrick3!1} 0.4\% & \cellcolor{Firebrick3!14} 4.1\% \\
 & Accuracy/Gender Mismatch & \cellcolor{Firebrick3!3} 0.8\% &  & \cellcolor{Firebrick3!5} 1.5\% & \cellcolor{Firebrick3!0} 0.0\% & \cellcolor{Firebrick3!2} 0.7\% & \cellcolor{Firebrick3!3} 0.8\% \\
 & Accuracy/Addition & \cellcolor{Firebrick3!5} 1.4\% &  & \cellcolor{Firebrick3!4} 1.1\% & \cellcolor{Firebrick3!6} 1.7\% & \cellcolor{Firebrick3!1} 0.4\% & \cellcolor{Firebrick3!8} 2.5\% \\
 & Style/Archaic or obscure word choice & \cellcolor{Firebrick3!3} 1.0\% &  & \cellcolor{Firebrick3!1} 0.4\% & \cellcolor{Firebrick3!6} 1.7\% & \cellcolor{Firebrick3!5} 1.5\% & \cellcolor{Firebrick3!1} 0.4\% \\
 & Terminology/Inappropriate for context & \cellcolor{Firebrick3!1} 0.2\% &  & \cellcolor{Firebrick3!0} 0.0\% & \cellcolor{Firebrick3!1} 0.4\% & \cellcolor{Firebrick3!0} 0.0\% & \cellcolor{Firebrick3!1} 0.4\% \\
\bottomrule
\end{tabular}
\smallskip

\begin{tabular}{ll >{\raggedleft\arraybackslash}p{11mm} c >{\raggedleft\arraybackslash}p{11mm}>{\raggedleft\arraybackslash}p{11mm} >{\raggedleft\arraybackslash}p{11mm}>{\raggedleft\arraybackslash}p{11mm}}
\toprule
& \bf Error type & All && EnCs & EnDe & Gemini2.5 & Human \\
\midrule
\parbox[t]{2mm}{\multirow{8}{*}{\rotatebox[origin=c]{90}{\bf MT-breaker (seeded)}}} & Accuracy/Mistranslation & \cellcolor{Firebrick3!99} 29.8\% &  & \cellcolor{Firebrick3!100} 35.4\% & \cellcolor{Firebrick3!71} 21.3\% & \cellcolor{Firebrick3!100} 30.2\% & \cellcolor{Firebrick3!98} 29.4\% \\
 & Style/Unnatural or awkward & \cellcolor{Firebrick3!56} 16.9\% &  & \cellcolor{Firebrick3!72} 21.6\% & \cellcolor{Firebrick3!32} 9.7\% & \cellcolor{Firebrick3!64} 19.2\% & \cellcolor{Firebrick3!48} 14.4\% \\
 & Accuracy/Creative Reinterpretation & \cellcolor{Firebrick3!31} 9.2\% &  & \cellcolor{Firebrick3!34} 10.1\% & \cellcolor{Firebrick3!26} 7.9\% & \cellcolor{Firebrick3!29} 8.8\% & \cellcolor{Firebrick3!32} 9.7\% \\
 & Fluency/Grammar & \cellcolor{Firebrick3!11} 3.3\% &  & \cellcolor{Firebrick3!16} 4.9\% & \cellcolor{Firebrick3!2} 0.7\% & \cellcolor{Firebrick3!8} 2.5\% & \cellcolor{Firebrick3!14} 4.1\% \\
 & Fluency/Punctuation & \cellcolor{Firebrick3!18} 5.5\% &  & \cellcolor{Firebrick3!16} 4.7\% & \cellcolor{Firebrick3!22} 6.7\% & \cellcolor{Firebrick3!17} 5.1\% & \cellcolor{Firebrick3!20} 5.9\% \\
 & Style/Bad sentence structure & \cellcolor{Firebrick3!10} 3.0\% &  & \cellcolor{Firebrick3!15} 4.4\% & \cellcolor{Firebrick3!2} 0.7\% & \cellcolor{Firebrick3!10} 3.1\% & \cellcolor{Firebrick3!9} 2.8\% \\
 & Accuracy/Source language fragment & \cellcolor{Firebrick3!11} 3.4\% &  & \cellcolor{Firebrick3!4} 1.2\% & \cellcolor{Firebrick3!22} 6.7\% & \cellcolor{Firebrick3!13} 4.0\% & \cellcolor{Firebrick3!9} 2.8\% \\
 & Accuracy/Omission & \cellcolor{Firebrick3!5} 1.6\% &  & \cellcolor{Firebrick3!2} 0.7\% & \cellcolor{Firebrick3!10} 3.0\% & \cellcolor{Firebrick3!7} 2.0\% & \cellcolor{Firebrick3!4} 1.2\% \\
 & Fluency/Spelling & \cellcolor{Firebrick3!5} 1.6\% &  & \cellcolor{Firebrick3!6} 1.7\% & \cellcolor{Firebrick3!5} 1.5\% & \cellcolor{Firebrick3!0} 0.0\% & \cellcolor{Firebrick3!11} 3.4\% \\
 & Accuracy/Gender Mismatch & \cellcolor{Firebrick3!4} 1.2\% &  & \cellcolor{Firebrick3!7} 2.0\% & \cellcolor{Firebrick3!0} 0.0\% & \cellcolor{Firebrick3!4} 1.1\% & \cellcolor{Firebrick3!4} 1.2\% \\
 & Accuracy/Addition & \cellcolor{Firebrick3!2} 0.6\% &  & \cellcolor{Firebrick3!2} 0.7\% & \cellcolor{Firebrick3!1} 0.4\% & \cellcolor{Firebrick3!4} 1.1\% & \cellcolor{Firebrick3!0} 0.0\% \\
 & Style/Archaic or obscure word choice & \cellcolor{Firebrick3!2} 0.6\% &  & \cellcolor{Firebrick3!2} 0.5\% & \cellcolor{Firebrick3!2} 0.7\% & \cellcolor{Firebrick3!1} 0.3\% & \cellcolor{Firebrick3!3} 0.9\% \\
 & Terminology/Inappropriate for context & \cellcolor{Firebrick3!3} 0.9\% &  & \cellcolor{Firebrick3!2} 0.5\% & \cellcolor{Firebrick3!5} 1.5\% & \cellcolor{Firebrick3!5} 1.4\% & \cellcolor{Firebrick3!1} 0.3\% \\
\bottomrule
\end{tabular}
\smallskip

\begin{tabular}{ll >{\raggedleft\arraybackslash}p{11mm} c >{\raggedleft\arraybackslash}p{11mm}>{\raggedleft\arraybackslash}p{11mm} >{\raggedleft\arraybackslash}p{11mm}>{\raggedleft\arraybackslash}p{11mm}}
\toprule
& \bf Error type & All && EnCs & EnDe & Gemini2.5 & Human \\
\midrule
\parbox[t]{2mm}{\multirow{8}{*}{\rotatebox[origin=c]{90}{\bf MT-breaker (seedless)}}} & Accuracy/Mistranslation & \cellcolor{Firebrick3!100} 33.1\% &  & \cellcolor{Firebrick3!100} 35.7\% & \cellcolor{Firebrick3!75} 22.4\% & \cellcolor{Firebrick3!100} 39.0\% & \cellcolor{Firebrick3!81} 24.4\% \\
 & Style/Unnatural or awkward & \cellcolor{Firebrick3!60} 17.9\% &  & \cellcolor{Firebrick3!58} 17.4\% & \cellcolor{Firebrick3!67} 20.0\% & \cellcolor{Firebrick3!71} 21.2\% & \cellcolor{Firebrick3!44} 13.1\% \\
 & Accuracy/Creative Reinterpretation & \cellcolor{Firebrick3!43} 12.9\% &  & \cellcolor{Firebrick3!47} 14.0\% & \cellcolor{Firebrick3!27} 8.2\% & \cellcolor{Firebrick3!50} 15.1\% & \cellcolor{Firebrick3!32} 9.7\% \\
 & Fluency/Grammar & \cellcolor{Firebrick3!18} 5.3\% &  & \cellcolor{Firebrick3!22} 6.6\% & \cellcolor{Firebrick3!0} 0.0\% & \cellcolor{Firebrick3!17} 5.0\% & \cellcolor{Firebrick3!19} 5.7\% \\
 & Fluency/Punctuation & \cellcolor{Firebrick3!11} 3.2\% &  & \cellcolor{Firebrick3!11} 3.4\% & \cellcolor{Firebrick3!8} 2.4\% & \cellcolor{Firebrick3!5} 1.5\% & \cellcolor{Firebrick3!19} 5.7\% \\
 & Style/Bad sentence structure & \cellcolor{Firebrick3!4} 1.1\% &  & \cellcolor{Firebrick3!5} 1.4\% & \cellcolor{Firebrick3!0} 0.0\% & \cellcolor{Firebrick3!3} 0.8\% & \cellcolor{Firebrick3!6} 1.7\% \\
 & Accuracy/Source language fragment & \cellcolor{Firebrick3!5} 1.4\% &  & \cellcolor{Firebrick3!5} 1.4\% & \cellcolor{Firebrick3!4} 1.2\% & \cellcolor{Firebrick3!5} 1.5\% & \cellcolor{Firebrick3!4} 1.1\% \\
 & Accuracy/Omission & \cellcolor{Firebrick3!8} 2.3\% &  & \cellcolor{Firebrick3!5} 1.4\% & \cellcolor{Firebrick3!20} 5.9\% & \cellcolor{Firebrick3!4} 1.2\% & \cellcolor{Firebrick3!13} 4.0\% \\
 & Fluency/Spelling & \cellcolor{Firebrick3!7} 2.1\% &  & \cellcolor{Firebrick3!5} 1.4\% & \cellcolor{Firebrick3!16} 4.7\% & \cellcolor{Firebrick3!1} 0.4\% & \cellcolor{Firebrick3!15} 4.5\% \\
 & Accuracy/Gender Mismatch & \cellcolor{Firebrick3!5} 1.4\% &  & \cellcolor{Firebrick3!6} 1.7\% & \cellcolor{Firebrick3!0} 0.0\% & \cellcolor{Firebrick3!3} 0.8\% & \cellcolor{Firebrick3!8} 2.3\% \\
 & Accuracy/Addition & \cellcolor{Firebrick3!5} 1.4\% &  & \cellcolor{Firebrick3!5} 1.4\% & \cellcolor{Firebrick3!4} 1.2\% & \cellcolor{Firebrick3!4} 1.2\% & \cellcolor{Firebrick3!6} 1.7\% \\
 & Style/Archaic or obscure word choice & \cellcolor{Firebrick3!2} 0.7\% &  & \cellcolor{Firebrick3!2} 0.6\% & \cellcolor{Firebrick3!4} 1.2\% & \cellcolor{Firebrick3!1} 0.4\% & \cellcolor{Firebrick3!4} 1.1\% \\
 & Terminology/Inappropriate for context & \cellcolor{Firebrick3!1} 0.2\% &  & \cellcolor{Firebrick3!1} 0.3\% & \cellcolor{Firebrick3!0} 0.0\% & \cellcolor{Firebrick3!0} 0.0\% & \cellcolor{Firebrick3!2} 0.6\% \\
\bottomrule
\end{tabular}
\smallskip

\caption{Distribution of MQM error categories. Only categories with at least 10 overall occurences are shown. The percentages are normalized for each column in each table independently. Only segments where human translation exists are considered.}
\label{tab:50-mqm_breakdown}
\end{table*}

\end{document}